\documentclass[10pt,twocolumn,letterpaper]{article}

\usepackage[pagenumbers]{iccv} 

\usepackage{multirow} 

\definecolor{iccvblue}{rgb}{0.21,0.49,0.74}
\usepackage[pagebackref,breaklinks,colorlinks,allcolors=iccvblue]{hyperref}

\def\paperID{4714} 
\def\confName{ICCV}
\def\confYear{2025}

\title{Dynamic Dictionary Learning for Remote Sensing Image Segmentation}

\author{
Xuechao Zou$^{1,*}$, Yue Li$^{2,}$\thanks{Equal contribution}, Shun Zhang$^{2}$, Kai Li$^{3}$, \\ Shiying Wang$^{2}$, Pin Tao$^{2,3}$, Junliang Xing$^{3}$, Congyan Lang$^{1,}$\thanks{Corresponding author} \\
$^1$School of Computer Science and Technology, Beijing Jiaotong University, Beijing, China \\
$^2$Department of Computer Technology and Applications, Qinghai University, Xining, China \\
$^3$Department of Computer Science and Technology, Tsinghua University, Beijing, China \\
}

\usepackage{tikz}
\definecolor{low}{RGB}{185, 101, 71}
\definecolor{middle low}{RGB}{248, 202, 155}
\definecolor{middle}{RGB}{211, 232, 158}
\definecolor{middle high}{RGB}{138, 191, 104}
\definecolor{high}{RGB}{92, 144, 77}

\definecolor{clear sky}{RGB}{79,253,199}
\definecolor{thick cloud}{RGB}{77,2,115}
\definecolor{thin cloud}{RGB}{251,255,41}
\definecolor{cloud shadow}{RGB}{221,53,223}

\definecolor{loveda background}{RGB}{255, 255, 255}
\definecolor{loveda building}{RGB}{255, 0, 0}
\definecolor{loveda road}{RGB}{255, 255, 0}
\definecolor{loveda water}{RGB}{0, 0, 255}
\definecolor{loveda barren}{RGB}{159, 129, 183}
\definecolor{loveda forest}{RGB}{0, 255, 0}
\definecolor{loveda agriculture}{RGB}{255, 195, 128}

\definecolor{uavid clutter}{RGB}{0, 0, 0}
\definecolor{uavid building}{RGB}{128, 0, 0}
\definecolor{uavid road}{RGB}{128, 64, 128}
\definecolor{uavid tree}{RGB}{0, 128, 0}
\definecolor{uavid lowveg}{RGB}{128, 128, 0}
\definecolor{uavid moving_car}{RGB}{64, 0, 128}
\definecolor{uavid static_car}{RGB}{192, 0, 192}
\definecolor{uavid human}{RGB}{64, 64, 0}

\definecolor{potsdam ImSurf}{RGB}{255, 255, 255}
\definecolor{potsdam Building}{RGB}{0, 0, 255}
\definecolor{potsdam LowVeg}{RGB}{0, 255, 255}
\definecolor{potsdam Tree}{RGB}{0, 255, 0}
\definecolor{potsdam Car}{RGB}{255, 204, 0}
\definecolor{potsdam Clutter}{RGB}{255, 0, 0}

\begin{document}
\maketitle

\begin{abstract}
Remote sensing image segmentation faces persistent challenges in distinguishing morphologically similar categories and adapting to diverse scene variations. While existing methods rely on implicit representation learning paradigms, they often fail to dynamically adjust semantic embeddings according to contextual cues, leading to suboptimal performance in fine-grained scenarios such as cloud thickness differentiation. This work introduces a dynamic dictionary learning framework that explicitly models class ID embeddings through iterative refinement. The core contribution lies in a novel dictionary construction mechanism, where class-aware semantic embeddings are progressively updated via multi-stage alternating cross-attention querying between image features and dictionary embeddings. This process enables adaptive representation learning tailored to input-specific characteristics, effectively resolving ambiguities in intra-class heterogeneity and inter-class homogeneity. To further enhance discriminability, a contrastive constraint is applied to the dictionary space, ensuring compact intra-class distributions while maximizing inter-class separability. Extensive experiments across both coarse- and fine-grained datasets demonstrate consistent improvements over state-of-the-art methods, particularly in two online test benchmarks (LoveDA and UAVid). Code is available at \href{https://anonymous.4open.science/r/D2LS-8267/}{https://anonymous.4open.science/r/D2LS-8267/}.
\end{abstract}

\section{Introduction}\label{sec: intro}  

\begin{figure}[t]
  \centering
   \includegraphics[width=0.9\linewidth]{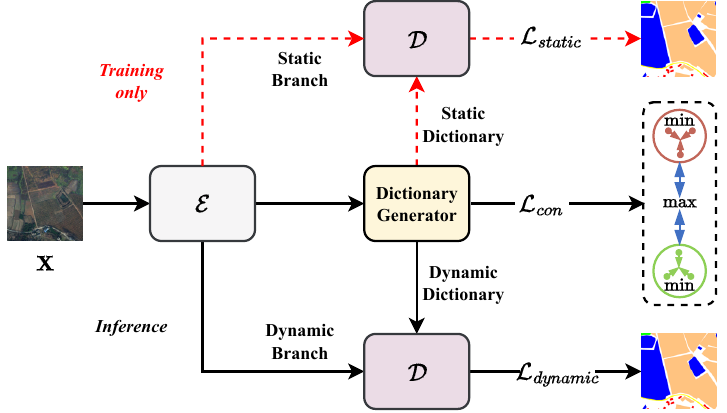}
   \caption{Overall pipeline of our dynamic dictionary learning. ``$\mathcal{E}$'' and ``$\mathcal{D}$'' denote the encoder and decoder, respectively.}
   \label{fig: pipeline}
   \vspace{-2.5ex}
\end{figure}

Remote sensing image segmentation aims to partition scenes into semantically consistent regions, supporting critical applications such as environmental monitoring~\cite{environmentalmonitoring}, disaster assessment~\cite{landslide}, and urban management~\cite{uvsam}. Traditional methods often fail to address the inherent complexity of remote sensing data, including fine-grained structural variations~\cite{ktda} (e.g., thin and thick clouds), intra-class heterogeneity (e.g., varying cloud textures), and inter-class homogeneity (e.g., farmland and grassland)~\cite{aerialformer}. 

Current research in remote sensing image segmentation has mainly explored three core directions. CNN-based methods like Hi-ResNet \cite{hiresnet} and CDNet \cite{cdnetv1,cdnetv2} prioritize edge optimization and lightweight design through multiscale architectures, while Transformer-based models such as DC-Swin \cite{dcswin} leverage global context via self-attention. Hybrid approaches (e.g., UNetFormer \cite{unetformer} and AerialFormer \cite{aerialformer}) integrate CNN-Transformer strengths for efficient multiscale feature fusion, addressing challenges like cloud segmentation \cite{hrcloudnet,mcdnet} and domain adaptation \cite{ktda}.

Despite these developments, existing methods still face notable issues when confronted with high intra-class heterogeneity and inter-class similarity in remote sensing images. Specifically, most approaches rely on implicit representation learning, which often fails to adapt to subtle morphological or textural differences, such as differentiating thin vs. thick clouds~\cite{ktda} or distinguishing visually similar urban objects~\cite{aerialformer}. Furthermore, while attention mechanisms~\cite{transformer,detr,vit, senet} capture global context, they do not inherently ensure discriminative class-specific feature representations, leading to confusion among closely related categories. This limitation becomes more pronounced in fine-grained segmentation tasks~\cite{cloud,ktda}, where even minor appearance variations within a class can significantly affect model performance. These shortcomings highlight the need to encode class-aware embeddings explicitly.

To overcome these limitations, we propose a dynamic dictionary learning framework that explicitly constructs and refines class ID embeddings through iterative interactions between image features and semantic dictionaries. Our key innovations are as follows:
\begin{itemize} 
\item We propose a dynamic dictionary learning framework that explicitly models class-aware semantic embeddings and leverages a contrastive loss to enhance intra-class compactness and inter-class separability, thereby improving performance, particularly in fine-grained remote sensing image segmentation tasks. 
\item We develop a modulator to convert the static dictionary into a dynamic one by generating attention maps from high-level semantic features, enabling input-specific adaptation and capturing subtle morphological variations within the same class. 
\item We design a multi-stage alternating cross-attention query decoder that progressively refines both image features and the semantic dictionary iteratively, facilitating mutual optimization and leading to more accurate and robust segmentation outcomes. 
\end{itemize}

Extensive experiments across both coarse-grained datasets (LoveDA~\cite{loveda}, UAVid~\cite{uavid}, Potsdam~\cite{potsdam}, Vaihingen~\cite{vaihingen}) and fine-grained segmentation tasks (Cloud~\cite{cloud}, Grass~\cite{ktda}) demonstrated consistent state-of-the-art performance. Our method exhibits superior generalization capabilities on two online test benchmarks (LoveDA and UAVid, where ground truth annotations are unavailable), further validating its robustness and segmentation effectiveness.

\section{Related Work}\label{sec: related_work}

\begin{figure*}[t]
  \centering
   \includegraphics[width=0.95\linewidth]{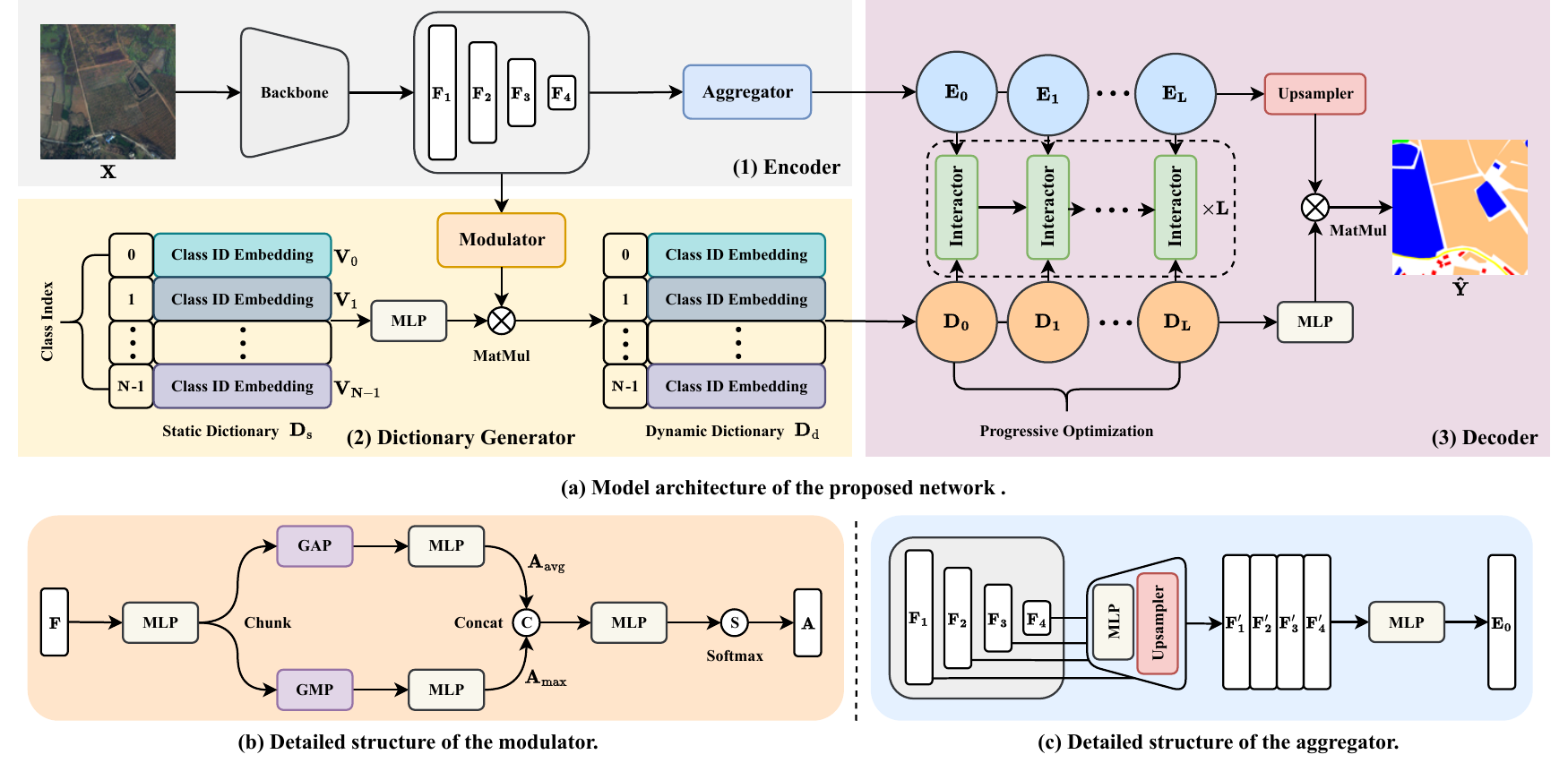}
   \caption{Overview of the proposed network architecture. Notably, we show only the dynamic branch for brevity.}
   \label{fig: network}
   \vspace{-2ex}
\end{figure*}

\subsection{Remote Sensing Image Segmentation}
Remote sensing image segmentation methods can be categorized as coarse- and fine-grained approaches. Coarse-grained segmentation focuses on large-scale semantic understanding of scenes, such as urban landscapes and land-cover classification. Recent advances have integrated Transformers and convolutional neural networks (CNNs) to capture global context and local details. For example, UNetFormer \cite{unetformer} employs a lightweight ResNet encoder and a Transformer-based decoder for real-time urban scene segmentation, while Hi-ResNet \cite{hiresnet} enhances edge details through multiscale branches and class-agnostic edge-aware loss. AerialFormer \cite{aerialformer} combined hierarchical Transformers with multi-dilated CNNs to address challenges like object density and intra-class heterogeneity. Other studies, such as BANet \cite{banet} and ABCNet \cite{abcnet}, further optimized feature fusion and computational efficiency through bilateral context modeling and lightweight architectures.

Fine-grained segmentation targets the precise delineation of objects such as thin and thick clouds. Cloud detection methods often leverage dual-branch architectures to balance spatial details and semantic context. DBNet \cite{dbnet} uses a Transformer-CNN hybrid with mutual guidance modules to refine cloud boundaries, and HRCloudNet \cite{hrcloudnet} integrates hierarchical high-resolution features for complex cloud textures. MCDNet \cite{mcdnet} addresses thin cloud detection via dual-perspective change-guided mechanisms. In contrast, lightweight models like SCNN \cite{scnn} and CDNet series \cite{cdnetv1, cdnetv2} reduce computational costs using shallow networks or attention-based refinement. Recent domain adaptation frameworks, such as KTDA \cite{ktda}, further enhance generalization by aligning features across domains. These methods collectively advance fine-grained segmentation accuracy. However, most existing methods struggle to distinguish morphologically similar classes, resulting in ambiguous predictions for fine-grained scenarios. To address this, we propose a dynamic dictionary learning framework that explicitly models class-aware embeddings, thereby enabling a more robust discrimination of related categories.

\subsection{Transformer and Attention Mechanisms}
Transformer-based approaches such as DETR~\cite{detr} reformulate object detection as a set prediction task, eliminating hand-crafted components like anchors and NMS. Subsequent works like DETR-style methods~\cite{dq-det,dynamic-detr,conditional-detr,focus-detr} optimize query mechanisms and attention patterns to accelerate convergence and improve accuracy. For segmentation, MaskFormer~\cite{maskformer,mask2former} and unify various segmentation tasks through mask classification, while SegViT~\cite{segvit,segvitv2} and SegFace~\cite{segface} leverage class tokens for semantic mask generation. These methods demonstrate the Transformer's versatility in dense prediction tasks.

Attention mechanisms~\cite{tdanet,pmaa,iianet,diffcr} play a pivotal role in enhancing feature discrimination and cross-modal interactions. Channel attention modules like SENet~\cite{senet}, CBAM~\cite{cbam}, and ECANet~\cite{ecanet} adaptively recalibrate channel-wise features, improving representation capacity. Cross-attention enables dynamic information exchange between heterogeneous data, as seen in CCNet~\cite{ccnet} for criss-cross context aggregation. Such mechanisms also drive vision foundational models like Segment Anything~\cite{segmentanything}, which utilizes prompt-based cross-attention for zero-shot segmentation. The synergy of diverse attention variants continues to push the boundaries of visual understanding systems. Inspired by DETR-style approaches and attention mechanisms, we explicitly model semantic labels with a learnable dictionary, further enabling progressive optimization via multi-stage alternating cross-attention, thereby effectively distinguishing morphologically similar classes in remote sensing images.

\section{Method}\label{sec: method}

\subsection{Overall Pipeline}
The overall pipeline for dynamic dictionary learning framework (as shown in ~\cref{fig: pipeline}) can be divided into two key parts: training and inference.

\subsubsection{Training}
During training, the pipeline employs a dual-branch architecture: the encoder $\mathcal{E}$ processes the input image $\mathbf{X} \in \mathbb{R}^{3 \times H \times W}$ and distributes the extracted features $\mathbf{F} \in \mathbb{R}^{C' \times H' \times W'}$ to two parallel branches: the static and dynamic branches. The static branch generates segmentation maps using a fixed static dictionary produced by a dictionary generator, while the dynamic branch operates with a continuously updated dynamic dictionary throughout training. Both branches share the same decoder. These processes can be formulated as:
\begin{equation}
    \mathbf{F} = \mathcal{E}(\mathbf{X}), \quad
    \mathbf{O_s}= \mathcal{D}(\mathbf{D_s}, \mathbf{F}), \quad \mathbf{O_d} = \mathcal{D}(\mathbf{D_d}, \mathbf{F}),
\end{equation}
where $\mathcal{E}(\cdot)$ and $\mathcal{D}(\cdot, \cdot)$ denote the encoder and decoder, respectively. $\mathbf{D_s} \in \mathbb{R}^{N \times C'} $ and $\mathbf{D_d} \in \mathbb{R}^{N \times C'}$ represent the static and dynamic dictionaries, while $\mathbf{O_\text{s}} \in \mathbb{R}^{N \times H \times W}$ and $\mathbf{O_\text{d}} \in \mathbb{R}^{N \times H \times W}$ are the segmentation outputs from the static and dynamic branches. $\mathbf{N}$ is the number of semantic categories, $H$, $W$, and $C'$ denote the height, width of the image, and the number of feature channels, respectively.


In the training process, we achieve joint optimization by minimizing the loss functions of the static and dynamic branches, namely \(\mathcal{L}_{\text{static}}\) and \(\mathcal{L}_{\text{dynamic}}\). The total loss function is defined as a weighted sum of the individual losses:
\begin{equation}
\mathcal{L}_{\text{total}} = \lambda_{\text{static}} \cdot \mathcal{L}_{\text{static}} + \lambda_{\text{dynamic}} \cdot \mathcal{L}_{\text{dynamic}},
\end{equation}
where \(\lambda_{\text{static}} = 0.4\) and \(\lambda_{\text{dynamic}} = 1\) are the default hyperparameters. During training, the network is optimized by minimizing the total loss \(\mathcal{L}_{\text{total}}\).

The loss for the static branch, \(\mathcal{L}_{\text{static}}\), is composed of a weighted sum of the cross-entropy loss and the Dice loss, which is a common design in segmentation tasks \cite{sfanet,segface}:
\begin{equation}
\mathcal{L}_{\text{static}} = \mathcal{L}_{\text{ce}}(\mathbf{O_\text{s}}, \mathbf{Y}) + \mathcal{L}_{\text{dice}}(\mathbf{O_\text{s}}, \mathbf{Y}),
\end{equation}
where \(\mathbf{O_\text{s}}\) denotes the segmentation prediction from the static branch, and \(\mathbf{Y} \in \mathbb{R}^{N \times H \times W}\) represents the ground truth segmentation label. The cross-entropy loss is employed for per-pixel classification, whereas the Dice loss focuses on the overlapping regions between the predicted and the ground truth segmentation maps, effectively mitigating the class imbalance problem, which is particularly important for segmentation tasks involving long-tail classes.

For the dynamic branch, the loss \(\mathcal{L}_{\text{dynamic}}\) not only includes the same loss terms as the static branch but also incorporates a dictionary-based contrastive loss \cite{clip} to ensure the embedding representations for each class are unique and discriminative. Specifically, the dynamic loss is defined as:
\begin{equation}
\mathcal{L}_{\text{dynamic}} = \mathcal{L}_{\text{ce}}(\mathbf{O_\text{d}}, \mathbf{Y}) + \mathcal{L}_{\text{dice}}(\mathbf{O_\text{d}}, \mathbf{Y}) + \mathcal{L}_{\text{con}}(\mathbf{D_\text{d}}).
\end{equation}

Moreover, the contrastive loss \(\mathcal{L}_{\text{con}}\) consists of an intra-class compactness loss \(\mathcal{L}_{\text{intra}}\) and an inter-class dispersion loss \(\mathcal{L}_{\text{inter}}\):
\begin{equation}
\mathcal{L}_{\text{con}} = \frac{\mathcal{L}_{\text{intra}}}{\mathcal{L}_{\text{inter}} + \epsilon},
\end{equation}
where \(\epsilon = 10^{-6}\) is used to prevent division by zero. The intra-class compactness loss measures the average Euclidean distance between the samples and their corresponding class centers:
\begin{equation}
\mathcal{L}_{\text{intra}} = \frac{1}{B \cdot N} \sum_{i=1}^N \sum_{b=1}^B \|\mathbf{D_d}^{(b,i)} - \boldsymbol{\mu}_i\|_2^2,
\end{equation}
where \(B\) denotes the batch size, \(N\) is the total number of classes, and \(\mathbf{D_\text{d}^{(b,i)}}\) represents the embedding of the \(b\)-th sample in class \(i\). The class center \(\boldsymbol{\mu}_i\) is computed as:
\begin{equation}
\boldsymbol{\mu}_i = \frac{1}{B} \sum_{b=1}^B \mathbf{D}_\text{d}^{(b,i)}.
\end{equation}
The inter-class dispersion loss aims to maximize the pairwise Euclidean distance between different class centers to enhance class separability:
\begin{equation}
\mathcal{L}_{\text{inter}} = \frac{2}{N(N-1)} \sum_{i=1}^{N-1} \sum_{j=i+1}^N \|\boldsymbol{\mu}_i - \boldsymbol{\mu}_j\|_2^2.
\end{equation}
The double summation iterates over all possible pairs of classes, ensuring adequate separation between them.

\subsubsection{Inference}
The network uses only the dynamic branch during inference, which utilizes the dynamic dictionary learned during training. The encoder processes the input image and generates feature maps, which are passed to the dynamic branch. The decoder then generates the segmentation maps. The dynamic dictionary complements the static dictionary by adaptively capturing the input-specific patterns, thereby improving the model’s generalization to unseen data.

\subsection{Proposed Network}
The proposed network architecture (see ~\cref{fig: network}) consists of an encoder, a dictionary generator, and a decoder, each playing a critical role in extracting features, learning dynamic dictionaries, and generating accurate segmentation maps.

\subsubsection{Encoder}
The encoder (as shown in \cref{fig: network}a1) is responsible for extracting feature representations from the input image. Initially, the input image $\mathbf{X}$ is processed by a backbone, generating multi-scale feature maps $\{ \mathbf{F}_i \in \mathbb{R}^{C_i \times H_i \times W_i} \}_{i=1}^4$, where $W_i=\frac{W}{2^i}$, $H_i=\frac{H}{2^i}$, and $C_i=C \times 2^i$. These feature maps are then passed through the aggregator, which performs feature mapping (\cref{equ: mapping}) and aggregation (\cref{equ: fusion}). The feature mapping step involves channel transformations, implemented by a multi-layer perceptron (MLP), and upsample interpolation to align all feature maps to the same dimensionality. After the mapping and interpolation, an MLP aggregates the features. The final output from the encoder is passed to the dictionary generator and the decoder.

\begin{equation}\label{equ: mapping}
\mathbf{F}_i' = \phi(\text{MLP}(\mathbf{F}_i)), \quad i \in \{1, 2, 3, 4\}
\end{equation}

\begin{equation}\label{equ: fusion}
\mathbf{E}_\text{0} = \text{MLP}(\mathbf{F}_1' \parallel \mathbf{F}_2' \parallel \mathbf{F}_3' \parallel \mathbf{F}_4'),    
\end{equation}
where $\{ \mathbf{F'}_i \in \mathbb{R}^{C' \times H' \times W'} \}_{i=1}^4$ are the feature maps after feature mapping, and $H'=H_1, W'=W_1$. \( \phi(\cdot) \) denotes the interpolation function. ``\(\parallel\)'' represents the concatenation operation along the channel dimension, and \( \mathbf{E}_0 \in \mathbb{R}^{C' \times H' \times W'}  \) is the final aggregated feature map.

\subsubsection{Dictionary Generator}
The dictionary generator is the core component of the proposed network and is responsible for creating static and dynamic dictionaries used in segmentation. These dictionaries help map feature information to class identifiers, enabling the model to distinguish different classes. The dynamic dictionary adapts during inference by modulating the static dictionary with input-specific attention. Unlike its static counterpart, it adjusts class ID embeddings to capture morphological variations across images, enhancing representation for nuanced inter-class distinctions.

\paragraph{Static Dictionary}
The static dictionary is a set of class ID embeddings used during the early stages of training. Each class ID embedding is associated with a learnable vector within the static dictionary. Once trained, these vectors are kept constant and do not change with input images, making them ``static." As a result, the static dictionary remains unchanged during inference. The static dictionary can be represented as:

\begin{equation}
\mathbf{D}_\text{s} = \{0: \mathbf{V}_0, 1: \mathbf{V}_1, \dots, \mathbf{N}-1: \mathbf{V}_{\mathbf{N}-1}\},
\end{equation}
where \( \mathbf{D_\text{s}} \in \mathbb{R}^{N \times C'} \) denotes the static dictionary, class index serves as the key, and \(\mathbf{V}_i \in \mathbb{R}^{C'}\) represents the embedding vector corresponding to the \( i \)-th class, which is the value in the dictionary. \( \mathbf{N} \) is the number of classes.

\paragraph{Dynamic Dictionary} 
The modulator (\cref{fig: network}b) produces an attention map from encoder features to convert the static dictionary into a dynamic dictionary. The feature map \( \mathbf{F} \in \mathbb{R}^{C_i \times H_i \times W_i} \) extracted from the encoder is first processed through an MLP layer. Specifically, we define \( \mathbf{F} = \mathbf{F_4} \), which encapsulates high-level semantic information. Subsequently, \( \mathbf{F} \) is split along the channel dimension into two separate chunks. One chunk undergoes global average pooling (GAP) followed by an MLP, while the other is processed through global max pooling (GMP) and a separate MLP. The outputs from these two branches are then concatenated along the channel dimension and fed into another MLP. Finally, the resulting feature representation is normalized using the softmax function, generating the attention map.
 
\begin{equation}
\mathbf{A}_\text{avg} = \text{MLP}(\text{GAP}(\mathbf{F}_\text{avg})),
\mathbf{A}_\text{max} = \text{MLP}(\text{GMP}(\mathbf{F}_\text{max})),
\end{equation}
where \( \mathbf{F}_\text{avg} \in \mathbb{R}^{\frac{C_i}{2} \times H_i \times W_i} \) and \( \mathbf{F}_\text{max} \in \mathbb{R}^{\frac{C_i}{2} \times H_i \times W_i} \) represent the two chunks of the feature map \( \mathbf{F} \). After concatenating the results of \( \mathbf{A}_\text{avg} \in \mathbb{R}^{\frac{C_i}{2}} \) and \( \mathbf{A}_\text{max} \in \mathbb{R}^{\frac{C_i}{2}} \), the final attention map \( \mathbf{A} \) is computed as:

\begin{equation}
\mathbf{A} = \text{Softmax}(\text{MLP}(\mathbf{A}_\text{avg} \parallel \mathbf{A}_\text{max})),
\end{equation}
where \( \mathbf{A} \in \mathbb{R}^{N \times r}  \) denotes the attention map, and $r=4$ represents the reduced dimensionality. The static dictionary \( \mathbf{D}_\text{s} \) is then transformed into the dynamic dictionary \( \mathbf{D}_\text{0} \in \mathbb{R}^{N \times C'} \) by matrix multiplication with the attention map \( \mathbf{A} \):

\begin{equation}
\mathbf{D}_\text{0} = \text{MLP}(\mathbf{D}_\text{s}) \otimes \mathbf{A},
\end{equation}
where \( \mathbf{D}_0 \) denotes the initial dynamic dictionary, \( \otimes \) denotes matrix multiplication. The initial dynamic dictionary is then passed to the decoder for iterative interaction and updates.

\subsubsection{Decoder}
In the decoder, the dynamic dictionary $ \mathbf{D}_0 $ and the image feature map $ \mathbf{E}_0 $ interact over $ \mathbf{L} $ stages for progressive optimization. During each stage, the dynamic dictionary and image feature alternately perform cross-attention queries, allowing for updates for both. The iterative process is:

\begin{equation}
\begin{aligned}
& \mathbf{D}_l \in \mathbb{R}^{N \times C'} = \text{softmax}\left( \frac{\mathbf{D}_{l-1} \mathbf{E}_{l-1}^\top}{\sqrt{d_k}} \right) \mathbf{E}_{l-1}, && \scriptstyle{\text{(D queries E)}} \\
& \mathbf{E}_l \in \mathbb{R}^{C' \times H' \times W'} = \text{softmax}\left( \frac{\mathbf{E}_{l-1} \mathbf{D}_l^\top}{\sqrt{d_k}} \right) \mathbf{D}_l, && \scriptstyle{\text{(E queries D)}}
\end{aligned}
\label{eq:alternating_attention}
\end{equation}
where $ l \in [1, \mathbf{L}] $ denotes the optimization stage. This alternating mechanism enables mutual enhancement between the dynamic dictionary and image features. Once the final feature map $ \mathbf{E}_\mathbf{L} $ is obtained after $ \mathbf{L} $ stages, the feature map is upsampled. Matrix multiplication is applied between the upsampled feature map and the dictionary $ \mathbf{D}_\mathbf{L} $ to produce the final segmentation output:

\begin{equation}
\mathbf{\hat{Y}} = \phi(\mathbf{E}_\mathbf{L}) \otimes \text{MLP}(\mathbf{D}_\mathbf{L}),
\end{equation}
where \( \mathbf{\hat{Y}} \in \mathbb{R}^{N\times H \times W} \) represents the final output.

\section{Experiments}\label{sec: experiments}

\subsection{Datasets}
We evaluated our method on six remote sensing datasets for semantic segmentation, spanning a range of altitudes from UAV to satellite imagery. These datasets include coarse-grained (LoveDA~\cite{loveda}, UAVid~\cite{uavid}, Potsdam~\cite{potsdam}, Vaihingen~\cite{vaihingen}) and fine-grained (Cloud~\cite{cloud}, Grass~\cite{ktda}) categories. For data splitting and preprocessing, the LoveDA, UAVid, Vaihingen, and Potsdam datasets follow the same protocol as in \cite{sfanet,unetformer}. In contrast, the Grass and Cloud datasets adopted preprocessing strategies identical to those used in~\cite{ktda}. LoveDA and UAVid are notably used in online test scenarios where ground truth labels are unavailable. For more details, please refer to \cref{sec: dataset_details}.

\subsection{Implementation Details}
\paragraph{Experimental Setup}
We set the size of the class ID embedding to 256 and the frequency of progressive optimization, denoted as $\mathbf{L}$, to 3. The backbone model was ConvNeXt-Base \cite{convnext}. The training configuration utilized a batch size of 4, with the random seed fixed at 42 to ensure reproducibility. The optimizer was AdamW \cite{adamw}, with an initial learning rate set to $1\times10^{-4}$ and weight decay set to 0.01. The learning rate was dynamically adjusted using a cosine annealing scheduler. All experiments were conducted on a workstation equipped with NVIDIA A100. 
\paragraph{Evaluation Metrics}
We followed the evaluation protocols of the comparative methods~\cite{sfanet,aerialformer}: Overall Accuracy (OA), intersection over union (IoU), F1 score, mean IoU (mIoU), and mean F1 (mF1). mIoU and mF1 are the averages of IoU and F1 scores across all classes. For readability, all metrics are presented in percentage form. Metric details and their formal definitions are provided in~\cref{sec: evaluation_metrics}.

\subsection{Ablation Study}

\paragraph{Effectiveness of Explicit Class Modeling} 
As quantified in ~\cref{tab:dictionary_comparison}, our analysis reveals that implicit representation learning (by removing the dictionary generator) leads to consistent performance degradation (mIoU) across all datasets. The particularly pronounced impact on fine-grained segmentation tasks substantiates the effectiveness of explicit class modeling in the framework. Furthermore, dditional ablation studies on the dimensionality of class ID embeddings and the backbone are available in \cref{sec: ablation_study}.

\paragraph{Static Branch or Dynamic Branch}
\cref{tab:ablation_study} revealed the complementary roles of loss function in different branches. Using only static loss (static branch) yields limited performance while adding dynamic loss (dynamic branch) improves segmentation performance. The contrastive loss provides critical discrimination power - its inclusion boosts Grass performance by +0.43, with LoveDA/Vaihingen also peaking at 55.27/85.27. These results consistently indicate that dynamic branch is superior to static branch, and that joint optimization of dual branches with the addition of contrastive loss can further enhance segmentation performance. 

\begin{table}[h]
\centering
\setlength{\tabcolsep}{1.2mm}
\caption{Ablation study on loss function of different branches.}
\label{tab:ablation_study}
\begin{tabular}{l|ccc}
\toprule
\multirow{2}{*}{\textbf{Method}} & \multicolumn{3}{c}{\textbf{mIoU}~$\uparrow$} \\
\cmidrule(lr){2-4}
 & LoveDA & Vaihingen & Grass \\
\midrule
$\mathcal{L_\text{static}}$ & 54.80 & 85.10 & 50.96 \\
$\mathcal{L_\text{dynamic}}$ & 54.93 & 85.17 & 51.38 \\
$\mathcal{L_\text{static}}$ + $\mathcal{L_\text{dynamic}}$ (w/o $\mathcal{L_\text{con}}$) & 54.95 & 85.17 & 51.53 \\
$\mathcal{L_\text{static}}$ + $\mathcal{L_\text{dynamic}}$ (w/ $\mathcal{L_\text{con}}$) & \textbf{55.27} & \textbf{85.27} & \textbf{51.96} \\
\bottomrule
\end{tabular}
\end{table}

\begin{figure*}[t]
  \centering
   \includegraphics[width=0.85\linewidth]{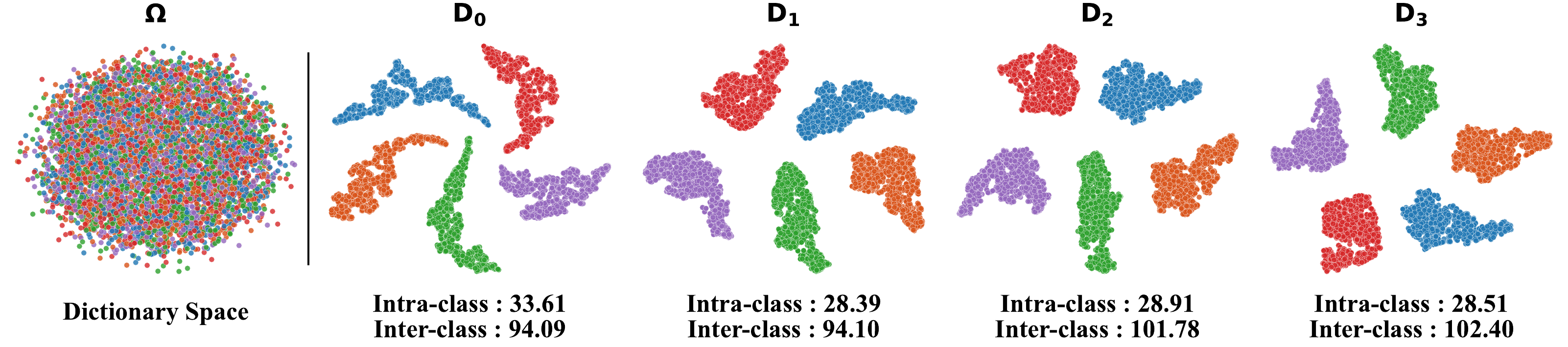}
   \caption{t-SNE visualization of dictionary distributions across different interaction stages.}
   \label{fig: tsne}
   \vspace{-2ex}
\end{figure*}

\paragraph{Interaction Stages}  
The feature-dictionary interaction stages progressively update class-aware semantic embeddings through multi-stage alternating cross-attention between image features and the dictionary. Experiments in~\cref{tab:depth_comparison} showed dataset-specific preferences: $\mathbf{L}=3$ yields optimal performance on most datasets, including LoveDA, UAVid, Grass and Cloud, while $\mathbf{L}=2$ suffices for Vaihingen (91.88 mF1). Excessive stages ($\mathbf{L}=4$) reduce segmentation performance on the Cloud dataset, indicating overfitting risks. Visualization of t-SNE (\cref{fig: tsne}) on the Grass dataset quantifies dictionary evolution: intra-class distances tighten ($\mathbf{D}_0$: 33.61 → $\mathbf{D}_3$: 28.51), while inter-class distances increase. We selected $\mathbf{L}=3$ as the default, balancing model performance and computational efficiency. 

\begin{table}[t]
\centering
\setlength{\tabcolsep}{0.9mm}
\caption{Performance comparison across interaction stages.}
\label{tab:depth_comparison}
\begin{tabular}{c|cccc
|cc}
\toprule
\multirow{2}{*}{$\mathbf{L}$} & \multicolumn{4}{c|}{\textbf{mIoU} $\uparrow$} & \multicolumn{2}{c}{\textbf{mF1} $\uparrow$} \\
\cmidrule(lr){2-5} \cmidrule(lr){6-7}
 & LoveDA & UAVid & Grass & Cloud & Potsdam & Vaihingen \\ 
\midrule
1 & 54.84 & 68.79 & 51.05 & 83.09 & 93.29 & 91.83 \\
2 & 55.00 & 69.49 & 51.21 & 82.95 & 94.09 & \textbf{91.88} \\
3 & \textbf{55.27} & \textbf{70.90} & \textbf{51.50} & \textbf{83.10} & \textbf{94.70} & 91.76 \\
4 & 54.84 & 67.88 & 50.90 & 80.97 & 93.87 & 91.85 \\
\bottomrule
\end{tabular}
\vspace{-2ex}
\end{table}

\vspace{-2ex}
\paragraph{Effectiveness of Each Component} 
We systematically validated the necessity of each component by disabling individual components. As shown in \cref{tab:module_comparison}, removing the modulator resulted in the most substantial decline in segmentation performance on the Vaihingen dataset, highlighting its critical role in handling complex urban scenes. Disabling the aggregator leads to the most significant performance drop on the Grass dataset, indicating its importance for fine-grained feature integration. The interactor contributed consistently across all datasets, demonstrating its capacity to refine ambiguous boundaries. The full model achieved peak performance, proving that these modules provide complementary functionalities rather than isolated improvements. This comprehensive ablation validates that all components are essential for segmentation performance.

\begin{table}[h]
\centering
\caption{Effectiveness of each component. Mod., Agg., Int. denote modulator, aggregator, and interaction respectively.}
\label{tab:module_comparison}
\begin{tabular}{ccc|ccc}
\toprule
\multirow{2}{*}{\textbf{Mod.}} & \multirow{2}{*}{\textbf{Agg.}} & \multirow{2}{*}{\textbf{Int.}} & \multicolumn{3}{c}{\textbf{mIoU}~$\uparrow$} \\
\cmidrule(lr){4-6}
 & & & LoveDA & Vaihingen & Grass \\
\midrule
$\times$ & $\checkmark$ & $\checkmark$ & 54.95 & 85.17 & 51.35 \\
$\checkmark$ & $\times$ & $\checkmark$ & 54.66 & 82.54 & 48.52 \\
$\checkmark$ & $\checkmark$ & $\times$ & 54.55 & 85.14 & 50.95 \\
$\checkmark$ & $\checkmark$ & $\checkmark$ & \textbf{55.27} & \textbf{85.27} & \textbf{51.96} \\
\bottomrule
\end{tabular}
\vspace{-2ex}
\end{table}

\begin{table*}[htbp]
\centering
\setlength{\tabcolsep}{2mm}
\caption{Comparison of semantic segmentation performance on the LoveDA dataset.}
\begin{tabular}{l|ccccccc|c}
\toprule
\textbf{Method} & \textbf{Background} & \textbf{Building} & \textbf{Road} & \textbf{Water} & \textbf{Barren} & \textbf{Forest} & \textbf{Agriculture} & \textbf{mIoU}~$\uparrow$ \\
\midrule
TransUNet~\cite{transunet} & 43.0 & 56.1 & 53.7 & 78.0 & 9.3 & 44.9 & 56.9 & 48.9 \\
DC-Swin~\cite{dcswin} & 41.3 & 54.5 & 56.2 & 78.1 & 14.5 & 47.2 & 62.4 & 50.6 \\
UNetFormer~\cite{unetformer} & 44.7 & 58.8 & 54.9 & 79.6 & 20.1 & 46.0 & 62.5 & 52.4 \\
Hi-Resnet~\cite{hiresnet} & 46.7 & 58.3 & 55.9 & 80.1 & 17.0 & 46.7 & 62.7 & 52.5 \\
AerialFormer~\cite{aerialformer} & \underline{47.8} & \underline{60.7} & \textbf{59.3} & 81.5 & 17.9 & \underline{47.9} & 64.0 & 54.1 \\
SFA-Net~\cite{sfanet} & \textbf{48.4} & 60.3 & 59.1 & \textbf{81.9} & \textbf{24.1} & 46.2 & \underline{64.0} & \underline{54.9} \\
Ours & 47.6 & \textbf{61.2} & \underline{59.1} & \underline{81.6} & \underline{23.8} & \textbf{48.8} & \textbf{64.8} & \textbf{55.3} \\
\bottomrule
\end{tabular}
\label{tab: loveda_results}
\end{table*}

\begin{table*}[htbp]
\centering
\setlength{\tabcolsep}{1.25mm}
\caption{Comparison of semantic segmentation performance on the UAVid dataset.}
\begin{tabular}{l|cccccccc|c}
\toprule
\textbf{Method} & \textbf{Clutter} & \textbf{Building} & \textbf{Road} & \textbf{Tree} & \textbf{Vegetaion} & \textbf{Moving Car} & \textbf{Static Car} & \textbf{Human} & \textbf{mIoU}~$\uparrow$ \\
\midrule
DANet~\cite{danet} & 64.9 & 58.9 & 77.9 & 68.3 & 61.5 & 59.6 & 47.4 & 9.1 & 60.6 \\
ABCNet~\cite{abcnet} & 67.4 & 86.4 & 81.2 & 79.9 & 63.1 & 69.8 & 48.4 & 13.9 & 63.8 \\
BANet~\cite{banet} & 66.7 & 85.4 & 80.7 & 78.9 & 62.1 & 69.3 & 52.8 & 21.0 & 64.6 \\
SegFormer~\cite{segformer} & 66.6 & 86.3 & 80.1 & 79.6 & 62.3 & 72.5 & 52.5 & 28.5 & 66.0 \\
UNetFormer~\cite{unetformer} & 68.4 & 87.4 & 81.5 & 80.2 & 63.5 & 73.6 & 56.4 & \underline{31.0} & 67.8 \\
SFA-Net~\cite{sfanet} & \underline{70.2} & \underline{89.0} & \underline{82.7} & \underline{80.8} & \underline{64.6} & \textbf{77.5} & \textbf{67.5} & 30.7 & \underline{70.4} \\
Ours & \textbf{71.0} & \textbf{89.7} & \textbf{83.2} & \textbf{82.1} & \textbf{66.1} & \underline{75.0} & \underline{59.0} & \textbf{41.4} & \textbf{70.9} \\
\bottomrule
\end{tabular}
\label{tab: uavid_results}
\end{table*}

\begin{table}[htbp]
\centering
\setlength{\tabcolsep}{1.25mm}
\caption{Comparison of semantic segmentation performance on the Potsdam and Vaihingen dataset.} 
\begin{tabular}{l|c|c|c}
\toprule
\multirow{2}{*}{\textbf{Method}} & \multicolumn{3}{c}{\textbf{mF1~$\uparrow$}} \\ 
\cmidrule(lr){2-4}
& \textbf{Potsdam} & \textbf{Vaihingen} & \textbf{Average} \\
\midrule
DANet~\cite{danet}        & 88.9   & 79.6   & 84.3 \\
ABCNet~\cite{abcnet}       & 92.7   & 89.5   & 91.1 \\
Segmenter~\cite{segmenter} & 89.2   & 84.1   & 86.7 \\
BANet~\cite{banet}         & 92.5   & 89.6   & 91.1 \\
SwinUperNet~\cite{swintransformer} & 92.2   & 89.8   & 91.0 \\
DC-Swin~\cite{dcswin}      & 93.3   & 90.7   & 92.0 \\
UNetFormer~\cite{unetformer} & 93.5   & 90.4   & 92.0 \\
AerialFormer~\cite{aerialformer} & \underline{94.1}   & 90.1   & 92.1 \\
SFA-Net~\cite{sfanet}      & 93.5   & \underline{91.2}   & \underline{92.4} \\
Ours                      & \textbf{94.7}   & \textbf{91.9}   & \textbf{93.3} \\
\bottomrule
\end{tabular}
\label{tab: potsdam_and_vaihingen_results}
\vspace{-3ex}
\end{table}

\subsection{Comparison with SOTA Methods}

\paragraph{Quantitative Analysis} Our method achieved consistent SOTA performance on six remote sensing semantic segmentation benchmarks. It also balances efficiency and performance well, with detailed computational efficiency comparisons (parameter count and FLOPs) in \textbf{Appendix D}.

\textbf{Coarse-grained Segmentation} 
Our method achieved state-of-the-art performance across all coarse-grained datasets. On the LoveDA dataset (\cref{tab: loveda_results}), we surpassed SFA-Net by +0.4\% mIoU, particularly in challenging categories like agriculture and building. The UAVid results (\cref{tab: uavid_results}) further validated our superiority in dynamic urban scenes, achieving 70.9 mIoU by learning distinct representations for different categories. For high-resolution aerial datasets, our approach established new benchmarks on Potsdam and Vaihingen (\cref{tab: potsdam_and_vaihingen_results}), with dominant performance in large-scale segmentation scenes. 

\textbf{Fine-grained Segmentation} 
As shown in ~\cref{tab: twofinegrained}, our method demonstrates remarkable advantages in distinguishing subtle class variations. On the Grass dataset, we achieve 66.27 mF1, with significant gains in differentiating coverage grades. For the Cloud dataset, our approach establishes an absolute lead, benefiting from the dynamic dictionary's adaptability to image feature variations. The 89.65 mF1 on Cloud highlights our capacity to resolve ambiguous boundaries through dynamic dictionary learning.
These results collectively confirm that our dynamic dictionary learning paradigm adapts effectively to large-scale scene variations and fine-grained semantic distinctions.

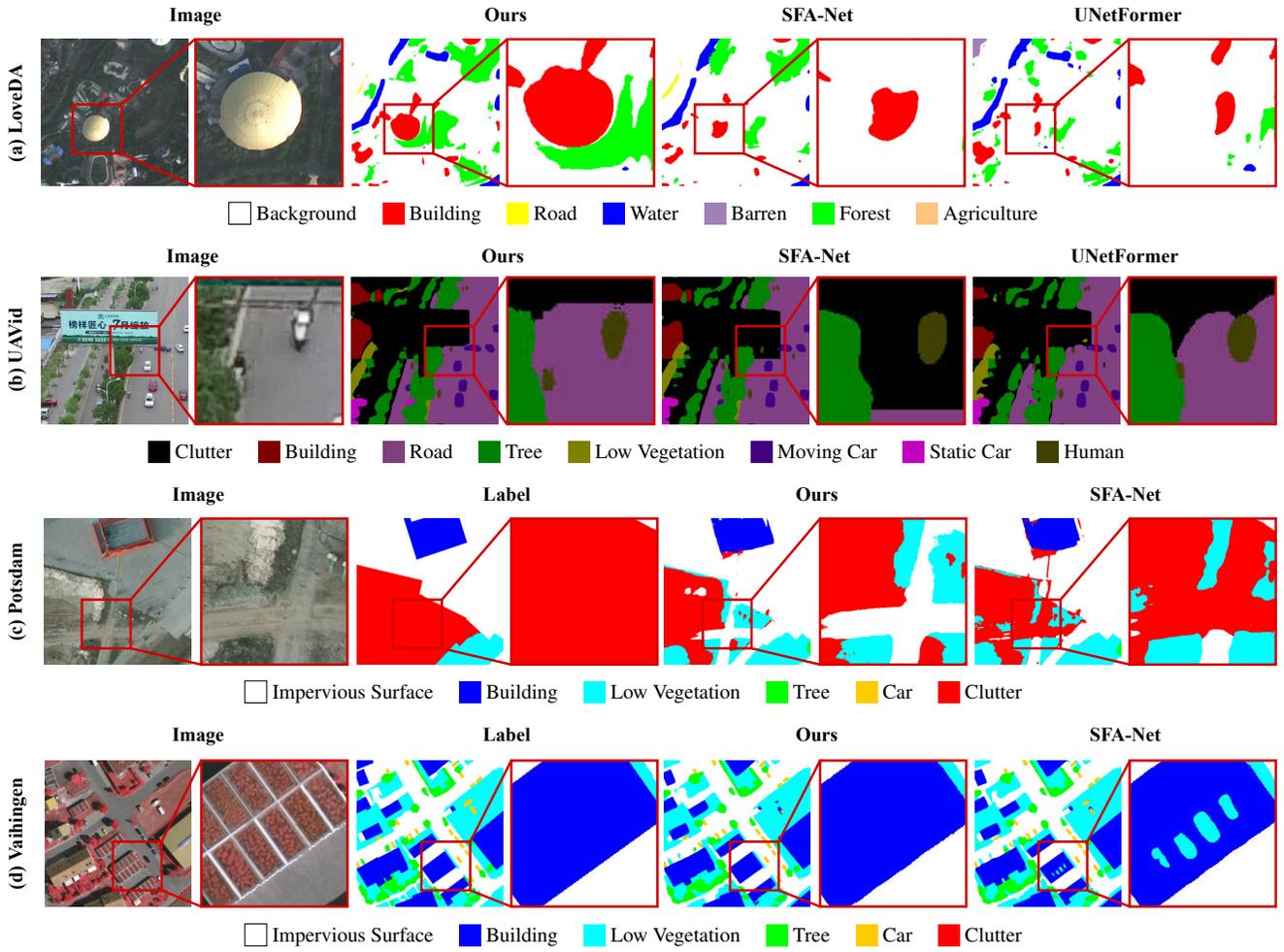
\begin{figure*}[htbp]
  \centering
  \begin{subfigure}[b]{\textwidth}
    \centering
    \includegraphics[width=\linewidth]{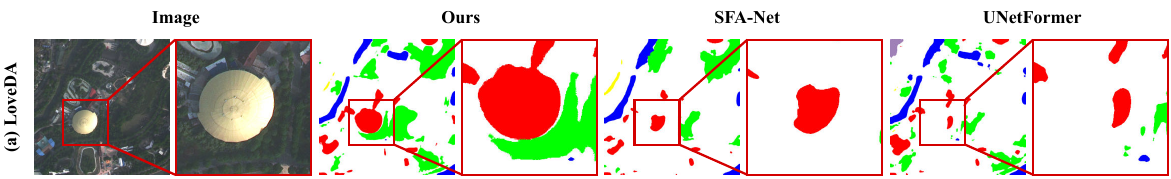}
    \begin{tikzpicture}
        \node {
            \fontsize{8}{8}\selectfont
            \begin{tabular}{*{7}{c@{\hspace{10pt}}}}
                \raisebox{-0.5ex}{\tikz\draw[black] (0,0) rectangle (0.3,0.3);} Background &
                \raisebox{-0.5ex}{\tikz\fill[loveda building] (0,0) rectangle (0.3,0.3);} Building &
                \raisebox{-0.5ex}{\tikz\fill[loveda road] (0,0) rectangle (0.3,0.3);} Road & 
                \raisebox{-0.5ex}{\tikz\fill[loveda water] (0,0) rectangle (0.3,0.3);} Water &
                \raisebox{-0.5ex}{\tikz\fill[loveda barren] (0,0) rectangle (0.3,0.3);} Barren &
                \raisebox{-0.5ex}{\tikz\fill[loveda forest] (0,0) rectangle (0.3,0.3);} Forest &
                \raisebox{-0.5ex}{\tikz\fill[loveda agriculture] (0,0) rectangle (0.3,0.3);} Agriculture
            \end{tabular}
        };
    \end{tikzpicture}
    \label{subfig:loveda}
  \end{subfigure}

  \begin{subfigure}[b]{\textwidth}
    \centering
    \includegraphics[width=\linewidth]{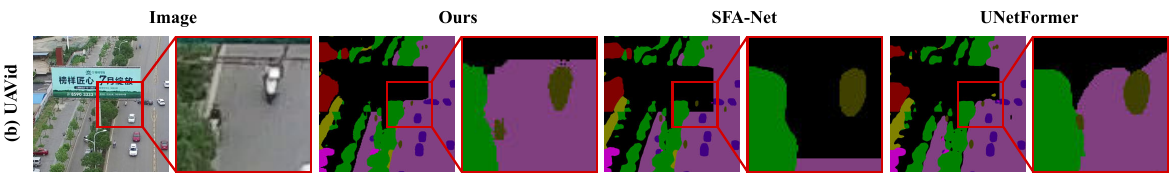}
    \begin{tikzpicture}
        \node {
            \fontsize{8}{8}\selectfont
            \begin{tabular}{*{8}{c@{\hspace{10pt}}}}
                \raisebox{-0.5ex}{\tikz\fill[uavid clutter] (0,0) rectangle (0.3,0.3);} Clutter &
                \raisebox{-0.5ex}{\tikz\fill[uavid building] (0,0) rectangle (0.3,0.3);} Building &
                \raisebox{-0.5ex}{\tikz\fill[uavid road] (0,0) rectangle (0.3,0.3);} Road &
                \raisebox{-0.5ex}{\tikz\fill[uavid tree] (0,0) rectangle (0.3,0.3);} Tree &
                \raisebox{-0.5ex}{\tikz\fill[uavid lowveg] (0,0) rectangle (0.3,0.3);} Low Vegetation &
                \raisebox{-0.5ex}{\tikz\fill[uavid moving_car] (0,0) rectangle (0.3,0.3);} Moving Car &
                \raisebox{-0.5ex}{\tikz\fill[uavid static_car] (0,0) rectangle (0.3,0.3);} Static Car &
                \raisebox{-0.5ex}{\tikz\fill[uavid human] (0,0) rectangle (0.3,0.3);} Human
            \end{tabular}
        };
    \end{tikzpicture}

    \label{subfig:uavid}
  \end{subfigure}

  \begin{subfigure}[b]{\textwidth}
    \centering
    \includegraphics[width=\linewidth]{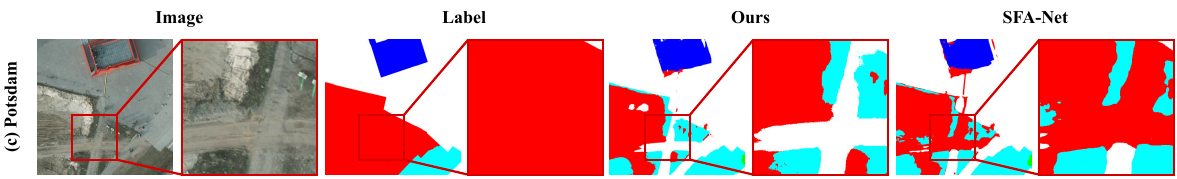}
    \begin{tikzpicture}
        \node {
            \fontsize{8}{8}\selectfont
            \begin{tabular}{*{6}{c@{\hspace{10pt}}}}
                \raisebox{-0.5ex}{\tikz\draw[black] (0,0) rectangle (0.3,0.3);} Impervious Surface &
                \raisebox{-0.5ex}{\tikz\fill[potsdam Building] (0,0) rectangle (0.3,0.3);} Building &
                \raisebox{-0.5ex}{\tikz\fill[potsdam LowVeg] (0,0) rectangle (0.3,0.3);} Low Vegetation &
                \raisebox{-0.5ex}{\tikz\fill[potsdam Tree] (0,0) rectangle (0.3,0.3);} Tree &
                \raisebox{-0.5ex}{\tikz\fill[potsdam Car] (0,0) rectangle (0.3,0.3);} Car &
                \raisebox{-0.5ex}{\tikz\fill[potsdam Clutter] (0,0) rectangle (0.3,0.3);} Clutter
            \end{tabular}
        };
    \end{tikzpicture}
    \label{subfig:potsdam1}
  \end{subfigure}

  \begin{subfigure}[b]{\textwidth}
    \centering
    \includegraphics[width=\linewidth]{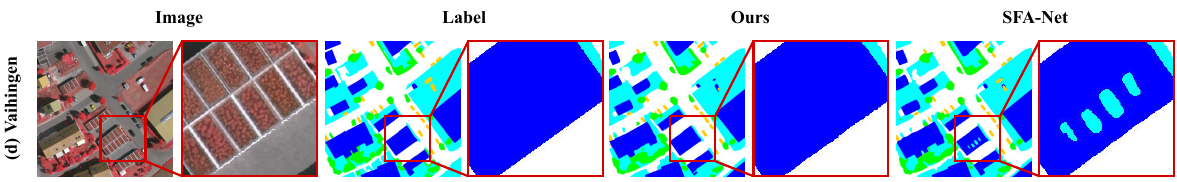}
    \begin{tikzpicture}
        \node {
            \fontsize{8}{8}\selectfont
            \begin{tabular}{*{6}{c@{\hspace{10pt}}}}
                \raisebox{-0.5ex}{\tikz\draw[black] (0,0) rectangle (0.3,0.3);} Impervious Surface &
                \raisebox{-0.5ex}{\tikz\fill[potsdam Building] (0,0) rectangle (0.3,0.3);} Building &
                \raisebox{-0.5ex}{\tikz\fill[potsdam LowVeg] (0,0) rectangle (0.3,0.3);} Low Vegetation &
                \raisebox{-0.5ex}{\tikz\fill[potsdam Tree] (0,0) rectangle (0.3,0.3);} Tree &
                \raisebox{-0.5ex}{\tikz\fill[potsdam Car] (0,0) rectangle (0.3,0.3);} Car &
                \raisebox{-0.5ex}{\tikz\fill[potsdam Clutter] (0,0) rectangle (0.3,0.3);} Clutter
            \end{tabular}
        };
    \end{tikzpicture}
    \label{subfig:potsdam2}
  \end{subfigure}

  \caption{Visualized results on four coarse-grained datasets. Ground truth labels are unavailable in the LoveDA and UAVid datasets.}
  \label{fig: all_results}
  \vspace{-1ex}
\end{figure*}

\begin{figure*}[htbp]
  \centering
  \begin{subfigure}[b]{\textwidth}
    \centering
    \includegraphics[width=\linewidth]{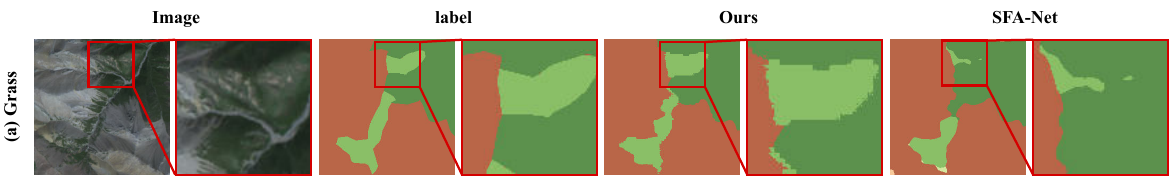}
    \begin{tikzpicture}
        \node {
            \fontsize{8}{8}\selectfont
            \begin{tabular}{*{5}{c@{\hspace{10pt}}}}
                \raisebox{-0.5ex}{\tikz\fill[low] (0,0) rectangle (0.3,0.3);} Low &
                \raisebox{-0.5ex}{\tikz\fill[middle low] (0,0) rectangle (0.3,0.3);} Middle-Low &
                \raisebox{-0.5ex}{\tikz\fill[middle] (0,0) rectangle (0.3,0.3);} Middle &
                \raisebox{-0.5ex}{\tikz\fill[middle high] (0,0) rectangle (0.3,0.3);} Middle-High &
                \raisebox{-0.5ex}{\tikz\fill[high] (0,0) rectangle (0.3,0.3);} High
            \end{tabular}
        };
    \end{tikzpicture}
    \label{subfig:new_legend1}
  \end{subfigure}

  \begin{subfigure}[b]{\textwidth}
    \centering
    \includegraphics[width=\linewidth]{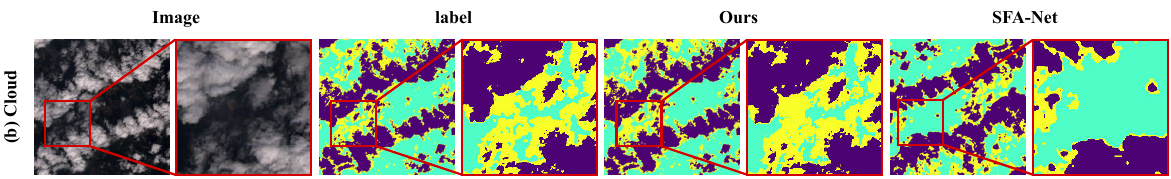}
    \begin{tikzpicture}
        \node {
            \fontsize{8}{8}\selectfont
            \begin{tabular}{*{4}{c@{\hspace{10pt}}}}
                \raisebox{-0.5ex}{\tikz\fill[clear sky] (0,0) rectangle (0.3,0.3);} Clear Sky &
                \raisebox{-0.5ex}{\tikz\fill[thick cloud] (0,0) rectangle (0.3,0.3);} Thick Cloud &
                \raisebox{-0.5ex}{\tikz\fill[thin cloud] (0,0) rectangle (0.3,0.3);} Thin Cloud &
                \raisebox{-0.5ex}{\tikz\fill[cloud shadow] (0,0) rectangle (0.3,0.3);} Cloud Shadow
            \end{tabular}
        };
    \end{tikzpicture}
    \label{subfig:new_legend2}
  \end{subfigure}
  
  \caption{Visualized results on two fine-grained datasets. More visualization of comparative methods can be found in Appendix E.}
  \label{fig: fine-grained-datasets}
  \vspace{3ex}
\end{figure*}
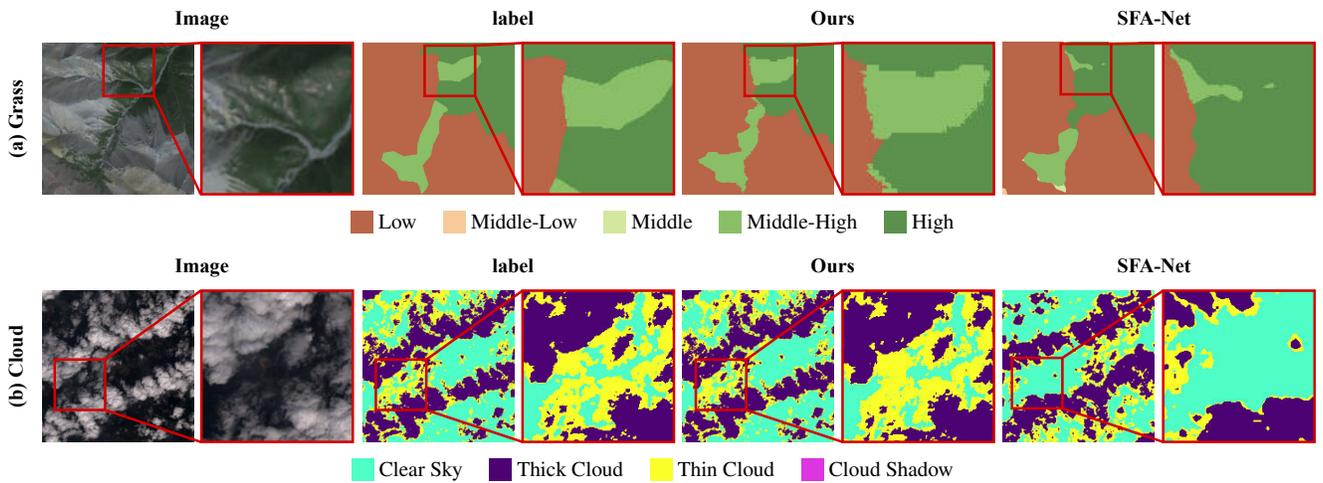

\paragraph{Visualization Results}  
As shown in \cref{fig: all_results}, our method accurately segments dome-shaped buildings in LoveDA, captures tiny human targets in UAVid without interference from vegetation and clutter, and correctly identifies rough road surfaces in Potsdam despite ground truth mislabeling. Additionally, it enables precise cloud-type separation in Cloud and enhances vegetation classification in Grass (\cref{fig: fine-grained-datasets}). More visualization results can be found in \cref{sec: more_visualization}. 

\begin{table}[htbp]
\centering
\setlength{\tabcolsep}{1mm}
\caption{Comparison of semantic segmentation performance on the two fine-grained datasets.}
\begin{tabular}{cc}
\begin{minipage}{0.23\textwidth}
\centering
\vspace{-1ex}
\caption*{(a) Grass}
\vspace{-1ex}
\begin{tabular}{l|c}  
\toprule
\textbf{Method} & \textbf{mF1}~$\uparrow$ \\ \midrule
FCN~\cite{fcn}         & 61.99          \\
PSPNet~\cite{pspnet}      & 62.55          \\
DeepLabV3+~\cite{deeplabv3+}  & 62.50          \\
UNet~\cite{unet}        & 62.34          \\
SegFormer~\cite{segformer}   & 62.82          \\
Mask2Former~\cite{mask2former} & 58.91          \\
DINOv2~\cite{dinov2}      & 61.70          \\ 
KTDA~\cite{ktda} & 65.01 \\
SFA-Net~\cite{sfanet} & \underline{65.76} \\
Ours        & \textbf{66.27} \\
\bottomrule
\end{tabular}
\end{minipage}
%
\hspace{-0.5mm}
\begin{minipage}{0.23\textwidth}
\centering
\vspace{-1ex}
\caption*{(b) Cloud}
\vspace{-1ex}
\begin{tabular}{l|c}
\toprule
\textbf{Method} & \textbf{mF1}~$\uparrow$ \\ \midrule
MCDNet~\cite{mcdnet}     & 42.76 \\
SCNN~\cite{scnn}       & 52.41 \\
CDNetv1~\cite{cdnetv1}    & 45.80  \\
KappaMask~\cite{kappamask}  & 68.47 \\
UNetMobv2~\cite{unetmobv2}  & 56.91 \\
CDNetv2~\cite{cdnetv2}    & 70.33 \\
HRCloudNet~\cite{hrcloudnet} & 71.36 \\ 
KTDA~\cite{ktda}       & 60.08 \\
SFA-Net~\cite{sfanet} & \underline{84.64} \\
Ours       & \textbf{89.65} \\
\bottomrule
\end{tabular}
\end{minipage}
\end{tabular}
\label{tab: twofinegrained}
\vspace{-2ex}
\end{table}

\section{Conclusion}
In this paper, we propose a dynamic dictionary learning framework to address the limitations of distinguishing similar categories in complex scenes caused by implicit representation learning in remote sensing image segmentation. Our method adaptively models morphological variations while preserving semantic discriminability by iteratively refining class-aware ID embeddings through alternating cross-attention querying and contrastive constraints. 

{
    \small
    \bibliographystyle{ieeenat_fullname}
    \bibliography{main}
}

\appendix
\maketitlesupplementary

\section{Dataset Details}~\label{sec: dataset_details}
As shown in~\cref{tab: all-datasets}, we provide comprehensive and detailed descriptions of all the datasets used in this paper.

\paragraph{LoveDA~\cite{loveda}}
The LoveDA dataset is a fine-resolution 0.3m dataset designed for urban and rural land cover classification. It consists of 5,987 images, each with a resolution of 512 × 512 pixels. Captured in three cities—Nanjing, Changzhou, and Wuhan, China—LoveDA spans both urban and rural environments. The metropolitan region contains dense infrastructure and complex geometries, while the rural area features natural landscapes and sparse settlements. This diversity in geographic environments provides valuable data for assessing the adaptability and generalizability of segmentation models. The dataset's varied land cover improves its versatility for real-world segmentation tasks.

\paragraph{UAVid~\cite{uavid}}
The UAVid dataset is a high-resolution dataset for semantic segmentation tasks in urban environments. It was captured by an unmanned aerial vehicle (UAV) flying at an altitude of 50 meters and consists of 42 video sequences and 420 images. The images are available in two spatial resolutions: 3,840 × 2,160 and 4,096 × 2,160 pixels. The dataset includes a diverse range of urban objects, such as buildings, roads, trees, vegetation, vehicles, humans, and other urban clutter. Both top-down and side views of urban scenes are provided, offering a comprehensive perspective for object recognition. In training stage, each image is divided into patches of 1,024 × 1,024 pixels.

\paragraph{Potsdam~\cite{potsdam}}
The ISPRS Potsdam dataset consists of 38 drone images from Potsdam, Germany, each with a resolution of 6000 × 6000 pixels and a ground sampling distance (GSD) of 5 cm, designed for semantic segmentation in urban environments. This dataset is annotated into six categories: impervious surfaces, buildings, low vegetation, trees, cars, and clutter. For the experiments, the images were cropped into 1,024 × 1,024-pixel patches to ensure manageable data processing and to focus on the primary RGB images and their corresponding labels.

\paragraph{Vaihingen~\cite{vaihingen}}
The ISPRS Vaihingen dataset includes 33 high-resolution images, each with a pixel resolution of 0.5 m, covering an urban region in Vaihingen, Germany. The images are classified into six categories: impervious surfaces, buildings, low vegetation, trees, cars, and clutter. The dataset’s average image size is 2,494 × 2,064 pixels.  For segmentation experiments, we used only the RGB images, and each image was divided into 1,024 × 1,024-pixel patches to facilitate efficient model training and testing.
\paragraph{Cloud~\cite{cloud}}
The Fine-Grained Cloud Segmentation dataset consists of 96 terrain-corrected (Level-1T) scenes from Landsat 8 OLI and TIRS, covering various biomes. This diverse dataset supports cloud detection and removal tasks in complex environments, offering pixel-level annotations for cloud shadow, clear sky, thin clouds, and cloud areas. Each scene is divided into 512 × 512-pixel patches and organized into training, validation, and test sets in a 6:2:2 ratio. The dataset’s wide range of cloud cover types and biomes makes it a valuable resource for training and evaluating segmentation models in cloud detection tasks.

\paragraph{Grass~\cite{ktda}}

This dataset was developed to overcome the limitations of existing grassland segmentation datasets, such as boundary ambiguity and misclassification in complex terrains. Created using high-resolution satellite imagery from Gaofen-2 and Gaofen-6, it was captured in 2019 over Maduo County, located in the Yellow River source area of China. The dataset includes high-resolution images (8m) and provides detailed grassland coverage classifications across five levels: low coverage, medium-low coverage, medium coverage, medium-high coverage, and high coverage. It consists of 1,500 pairs of 256 × 256-pixel patches, with manual refinement to ensure high accuracy. This dataset is especially valuable for fine-grained grassland extraction in high-altitude, ecologically sensitive regions.

\begin{table*}[t]
\setlength{\tabcolsep}{8pt}
\renewcommand{\arraystretch}{1.3}
\renewcommand{\arraystretch}{1.3}
\caption{Overview of remote sensing image datasets for semantic segmentation.}
\label{tab: all-datasets}
\begin{tabular}{cccccp{5cm}}
\toprule
\textbf{Type}                   & \textbf{Dataset} & \textbf{Source} & \textbf{GSD}\textsuperscript{*,†}  & \textbf{Patch Size} & \multicolumn{1}{c}{\textbf{Category}}                                               \\ \midrule
\multirow{4}{*}[-2.3em]{\centering Coarse Grained}
 & LoveDA~\cite{loveda}           & Satellite       & 0.3m          & 1024 * 1024         & background, building, road, water, barren, forest, agriculture  \\ 
                                & UAvid~\cite{uavid}            & UAV             & 50m\textsuperscript{†} & 1024 * 1024         & clutter, building, road, tree, low vegetation, moving car, static car, human \\  
                                & Potsdam~\cite{potsdam}          & UAV             & 5cm           & 1024 * 1024         & impervious surface, building, low vegetation, tree, car, clutter      \\ 
                                & Vaihingen~\cite{vaihingen}        & UAV             & 9cm          & 1024 * 1024         & impervious surface, building, low vegetation, tree, car, clutter      \\ \midrule
\multirow{2}{*}[-1.1em]{Fine Grained}   & Cloud~\cite{cloud}            & Satellite       & 30m           & 512 * 512           & cloud shadow, clear sky, thin cloud, thick cloud           \\  
                                & Grass~\cite{ktda}            & Satellite       & 8m            & 256 * 256           & low, medium-low, medium, medium-high, high                      \\ \bottomrule
\end{tabular}
\vspace{-3mm}
\begin{minipage}{\textwidth}
\footnotesize
\textsuperscript{*} \textit{GSD} (Ground Sampling Distance): The physical pixel size projected onto the ground surface (e.g., 0.3m/pixel = each pixel represents a 0.3×0.3m ground area). Smaller GSD indicates higher spatial resolution. \\
\textsuperscript{†} For UAVid dataset: The value indicates flight altitude rather than actual GSD. True GSD can be calculated via camera parameters.
\end{minipage}
\end{table*}

\section{Evaluation Metrics}~\label{sec: evaluation_metrics}

\paragraph{Overall Accuracy (OA)} calculates the ratio of correct predictions to the total number of pixels:
\begin{equation}
\text{OA} = \frac{\sum_{i} TP_i}{\sum_{i} (TP_i + FP_i + FN_i)},
\end{equation}
where \( TP_i \), \( FP_i \), and \( FN_i \) represent true positives, false positives, and false negatives for the \(i\)-th class, respectively.

\paragraph{Intersection over Union (IoU)} measures the overlap between predicted and ground truth regions for a class. Mean IoU (mIoU) averages the IoU scores across all classes.
\begin{equation}
\text{IoU} = \frac{TP}{TP + FP + FN},
\end{equation}

\begin{equation}
\text{mIoU} = \frac{1}{N} \sum_{i=1}^N \text{IoU}_i,
\end{equation}
where \( N \) is the number of classes, and \( \text{IoU}_i \) is the IoU for the \(i\)-th class.

\paragraph{F1} balances precision and recall, providing a harmonic mean of the two. Mean F1 score (mF1) averages the F1.
\begin{equation}
\text{F1} = 2 \cdot \frac{\text{Precision} \cdot \text{Recall}}{\text{Precision} + \text{Recall}},
\end{equation}

\begin{equation}
\text{mF1} = \frac{1}{N} \sum_{i=1}^N \text{F1}_i,
\end{equation}
where Precision \( = \frac{TP}{TP + FP} \) and Recall \( = \frac{TP}{TP + FN} \). \( \text{F1}_i \) balances precision and recall for the \(i\)-th class.

While OA provides a general performance measure, it is less reliable for imbalanced datasets. In contrast, \textbf{IoU} and \textbf{F1 score} (and their mean versions, \textbf{mIoU} and \textbf{mF1}) offer more robust evaluations by considering class-level performance, making them better suited for multi-class tasks.

\section{Ablation Study}~\label{sec: ablation_study}
\begin{figure}[t]
    \centering
    \includegraphics[width=\linewidth]{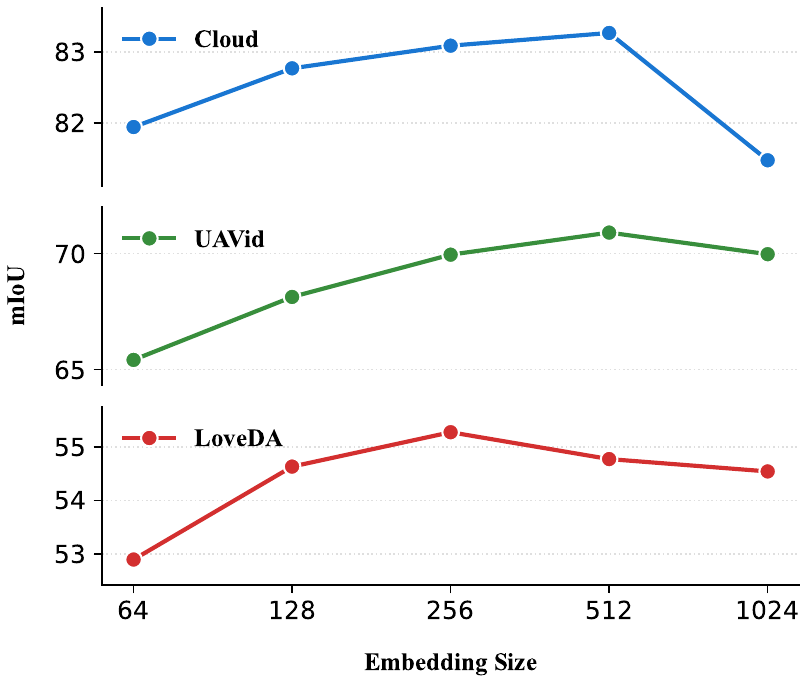}
    \caption{Performance comparison across different class ID embedding sizes for Cloud, UAVid, and LoveDA datasets. The mIoU scores are shown for embedding sizes ranging from 64 to 1024.}
    \label{fig: embedding_size_ablation}
\end{figure}

\paragraph{Dimensions of Class ID Embedding} 
We analyzed the impact of class ID embedding dimensions on segmentation performance. As shown in~\cref{fig: embedding_size_ablation}, increasing dimensions from 64 to 256 boosts performance across all datasets. While 512 dimensions achieve peak results on UAVid and Cloud, LoveDA slightly degrades, suggesting dimension sensitivity varies with dataset complexity. Extreme dimensions (1024) caused performance collapse, indicating overfitting risks. Therefore, we select 256 as the default value for achieving optimal efficiency-discriminability trade-off.

\paragraph{Backbone in Encoder}
We evaluate the impact of different backbone architectures on segmentation performance under frozen and fully trained settings using the LoveDA~\cite{loveda} dataset, as shown in~\cref{tab: ablation_on_backbone}. Among the tested backbones, Swin (B) achieves the highest mIoU of 49.76\% in the frozen setting, while ConvNeXt (B) outperforms others with 55.27\% when fully trained, demonstrating its superior feature extraction capability when fine-tuned. Swinv2 (B) and Swin (B) exhibit comparable performance, with Swinv2 (B) slightly behind in both settings. MobileNetv3 (L) and EfficientNet (M), being lightweight backbones, yield lower mIoU scores, particularly in the frozen setting, indicating their limited capacity to generalize without fine-tuning. These results suggest that transformer-based backbones generally perform better than CNN-based alternatives, and that full training significantly boosts segmentation performance across all architectures.

\begin{table}[t]
\centering
\setlength{\tabcolsep}{5mm}
\caption{Performance comparison of different backbones under frozen and fully-trained settings on the LoveDA dataset.}
\label{tab: ablation_on_backbone}
\begin{tabular}{llc}
\toprule
\textbf{Backbone} & \textbf{Status} & \textbf{mIoU~$\uparrow$} \\
\midrule
\multirow{2}{*}{ConvNeXt (B)~\cite{convnext}} & Frozen & 48.50 \\
                                  & Full    & 55.27 \\
\multirow{2}{*}{Swin (B)~\cite{swintransformer}}      & Frozen & 49.76 \\
                                  & Full    & 53.88 \\
\multirow{2}{*}{Swinv2 (B)~\cite{swinv2}}    & Frozen & 48.43 \\
                                  & Full    & 54.41 \\
\multirow{2}{*}{MobileNetv3 (L)~\cite{mobileNetV3}}       & Frozen & 46.34 \\
                                  & Full    & 51.17 \\
\multirow{2}{*}{EfficientNet (M)~\cite{efficientnetv2}}    & Frozen & 44.93 \\
                                  & Full    & 53.78 \\
\bottomrule
\end{tabular}
\end{table}

\section{Computational Efficiency}~\label{sec: computational_efficiency}

As shown in~\cref{tab: efficience}, our method achieves a substantial efficiency-performance tradeoff. Excluding the backbone, it has only 3.3M parameters, remaining lightweight. Though SFA-Net~\cite{sfanet} has fewer FLOPs, its minimal parameters limit representation capacity. Remarkably, our model achieves significant performance improvements with fewer parameters than both AerialFormer~\cite{aerialformer} and KTDA~\cite{ktda}.

\begin{table}[t]
\setlength{\tabcolsep}{1mm}
\caption{Computational efficiency comparison with input size 512$\times$512, including  \textsuperscript{*} indicates parameters without backbone.}
\label{tab: efficience}
\begin{tabular}{lcccc}
\toprule
Method & Params (M) & Params\textsuperscript{*} (M) & Flops (G) \\ \midrule
AerialFormer~\cite{aerialformer} & 113.8       & 26.9        & 126.8      \\
SFA-Net~\cite{sfanet} & \textbf{10.7}    & \textbf{0.6}     & \textbf{7.1}    \\
KTDA~\cite{ktda}    & 258.3       & 170.7       & 566.4      \\ 
Ours         & \underline{90.9}  & \underline{3.3}   & \underline{89.2}  \\
\bottomrule
\end{tabular}
\vspace{-2ex}
\end{table}

\section{More Visualization}~\label{sec: more_visualization}
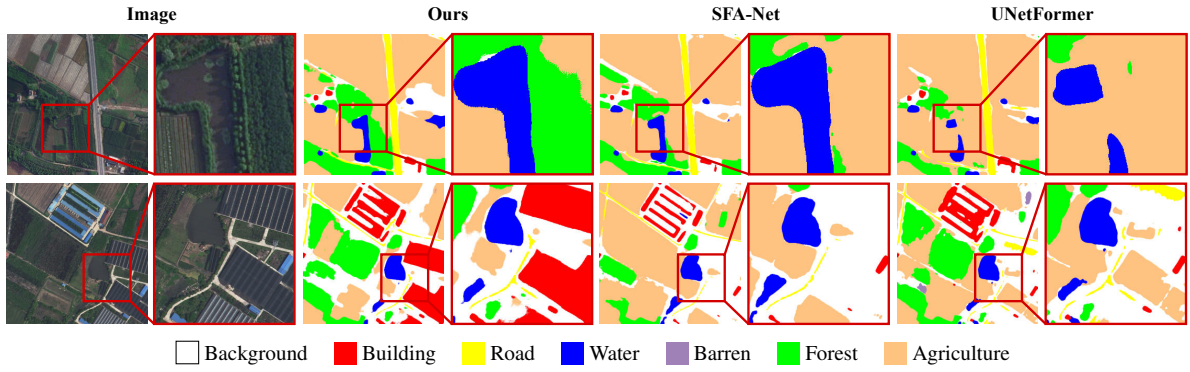
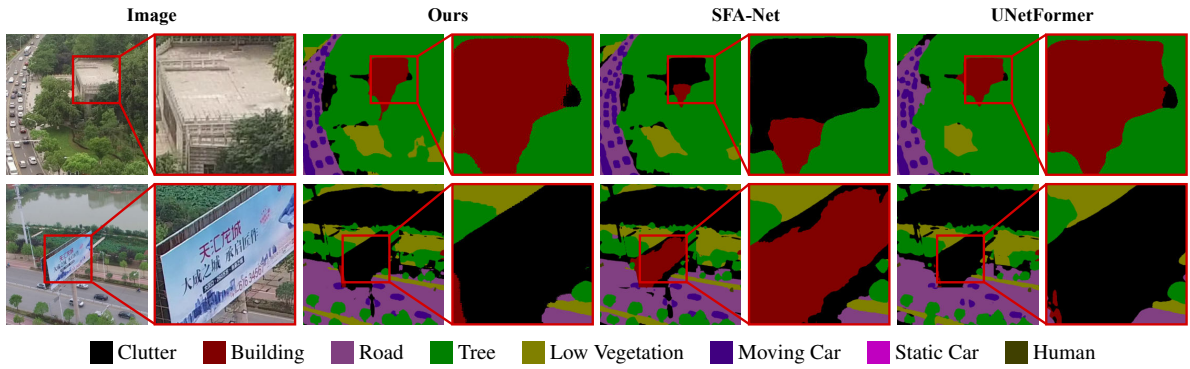
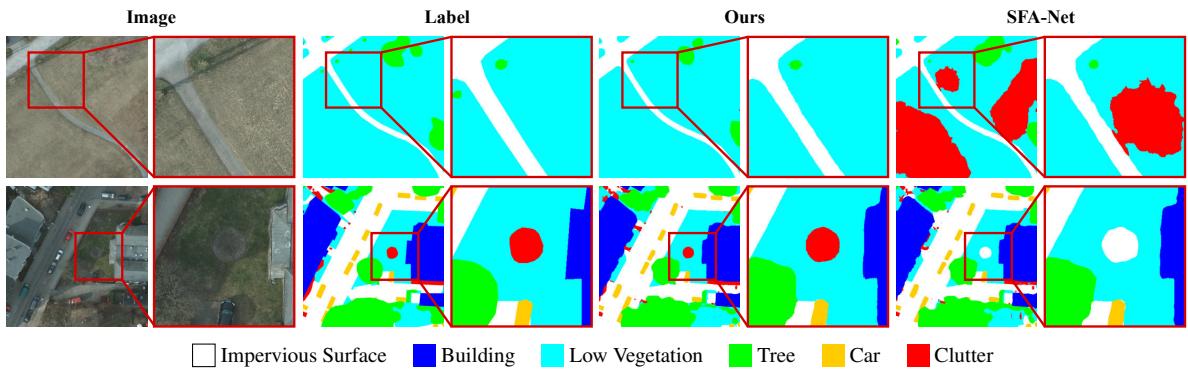
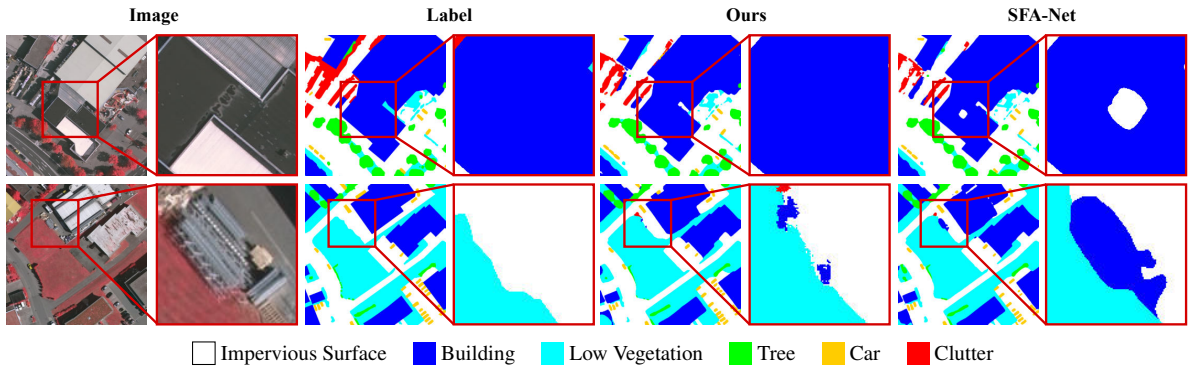
\begin{figure*}[t]
\centering
    \begin{subfigure}[t]{0.9\textwidth}
        \centering
        \includegraphics[width=\textwidth]{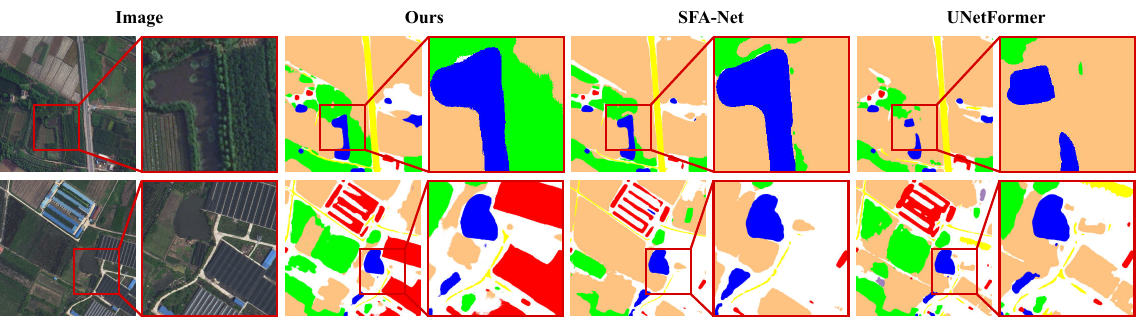}
        \begin{tikzpicture}
            \node {
                \fontsize{8}{8}\selectfont
                \begin{tabular}{*{7}{c@{\hspace{10pt}}}}
                    \raisebox{-0.5ex}{\tikz\draw[black] (0,0) rectangle (0.3,0.3);} Background &
                    \raisebox{-0.5ex}{\tikz\fill[loveda building] (0,0) rectangle (0.3,0.3);} Building &
                    \raisebox{-0.5ex}{\tikz\fill[loveda road] (0,0) rectangle (0.3,0.3);} Road & 
                    \raisebox{-0.5ex}{\tikz\fill[loveda water] (0,0) rectangle (0.3,0.3);} Water &
                    \raisebox{-0.5ex}{\tikz\fill[loveda barren] (0,0) rectangle (0.3,0.3);} Barren &
                    \raisebox{-0.5ex}{\tikz\fill[loveda forest] (0,0) rectangle (0.3,0.3);} Forest &
                    \raisebox{-0.5ex}{\tikz\fill[loveda agriculture] (0,0) rectangle (0.3,0.3);} Agriculture
                \end{tabular}
            };
        \end{tikzpicture}
        \caption{LoveDA}
        \label{fig: LoveDA}
    \end{subfigure}
    
    \begin{subfigure}[t]{0.9\textwidth}
        \centering
        \includegraphics[width=\textwidth]{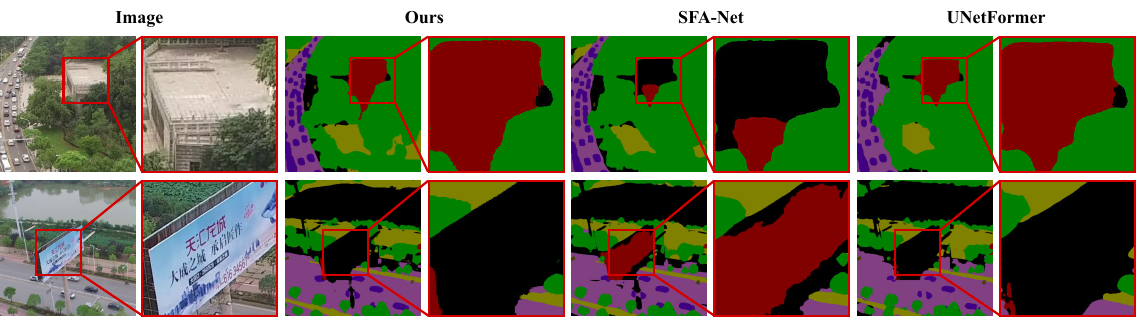}
        \begin{tikzpicture}
            \node {
                \fontsize{8}{8}\selectfont
                \begin{tabular}{*{8}{c@{\hspace{10pt}}}}
                    \raisebox{-0.5ex}{\tikz\fill[uavid clutter] (0,0) rectangle (0.3,0.3);} Clutter &
                    \raisebox{-0.5ex}{\tikz\fill[uavid building] (0,0) rectangle (0.3,0.3);} Building &
                    \raisebox{-0.5ex}{\tikz\fill[uavid road] (0,0) rectangle (0.3,0.3);} Road &
                    \raisebox{-0.5ex}{\tikz\fill[uavid tree] (0,0) rectangle (0.3,0.3);} Tree &
                    \raisebox{-0.5ex}{\tikz\fill[uavid lowveg] (0,0) rectangle (0.3,0.3);} Low Vegetation &
                    \raisebox{-0.5ex}{\tikz\fill[uavid moving_car] (0,0) rectangle (0.3,0.3);} Moving Car &
                    \raisebox{-0.5ex}{\tikz\fill[uavid static_car] (0,0) rectangle (0.3,0.3);} Static Car &
                    \raisebox{-0.5ex}{\tikz\fill[uavid human] (0,0) rectangle (0.3,0.3);} Human
                \end{tabular}
            };
        \end{tikzpicture}
        \caption{UAVid}
        \label{fig: UAVid}
    \end{subfigure}
    
    \begin{subfigure}[t]{0.9\textwidth}
        \centering
        \includegraphics[width=\textwidth]{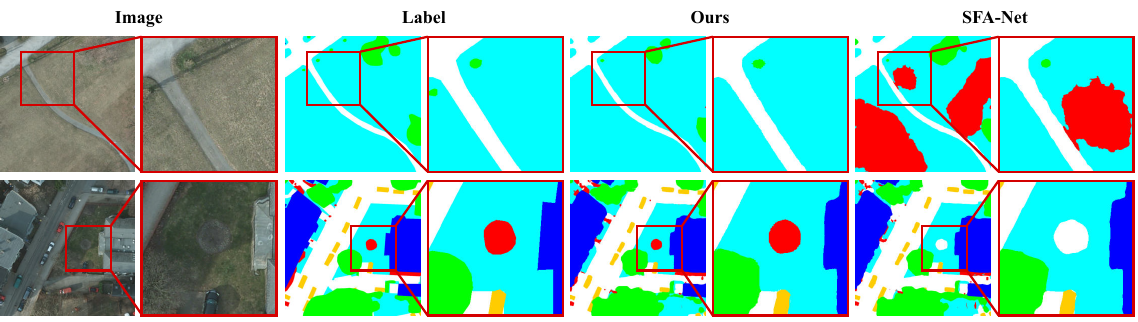}
        \begin{tikzpicture}
            \node {
                \fontsize{8}{8}\selectfont
                \begin{tabular}{*{6}{c@{\hspace{10pt}}}}
                    \raisebox{-0.5ex}{\tikz\draw[black] (0,0) rectangle (0.3,0.3);} Impervious Surface &
                    \raisebox{-0.5ex}{\tikz\fill[potsdam Building] (0,0) rectangle (0.3,0.3);} Building &
                    \raisebox{-0.5ex}{\tikz\fill[potsdam LowVeg] (0,0) rectangle (0.3,0.3);} Low Vegetation &
                    \raisebox{-0.5ex}{\tikz\fill[potsdam Tree] (0,0) rectangle (0.3,0.3);} Tree &
                    \raisebox{-0.5ex}{\tikz\fill[potsdam Car] (0,0) rectangle (0.3,0.3);} Car &
                    \raisebox{-0.5ex}{\tikz\fill[potsdam Clutter] (0,0) rectangle (0.3,0.3);} Clutter
                \end{tabular}
            };
        \end{tikzpicture}
        \caption{Potsdam}
        \label{fig: Potsdam}
    \end{subfigure}
    
    \begin{subfigure}[t]{0.9\textwidth}
        \centering
        \includegraphics[width=\textwidth]{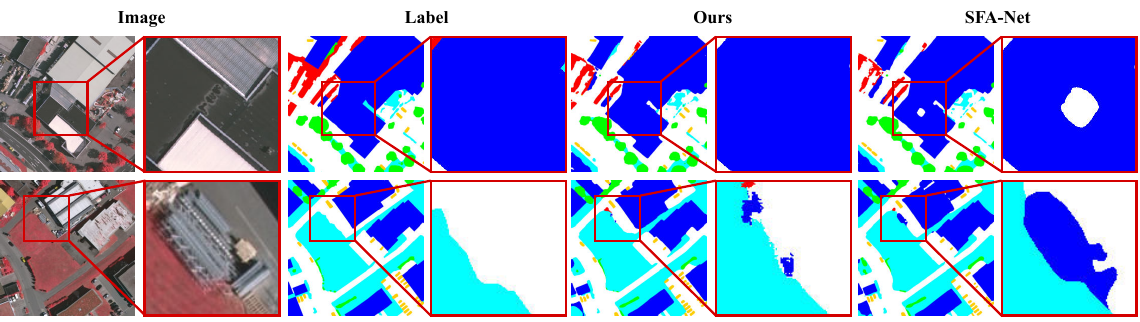}
        \begin{tikzpicture}
            \node {
                \fontsize{8}{8}\selectfont
                \begin{tabular}{*{6}{c@{\hspace{10pt}}}}
                    \raisebox{-0.5ex}{\tikz\draw[black] (0,0) rectangle (0.3,0.3);} Impervious Surface &
                    \raisebox{-0.5ex}{\tikz\fill[potsdam Building] (0,0) rectangle (0.3,0.3);} Building &
                    \raisebox{-0.5ex}{\tikz\fill[potsdam LowVeg] (0,0) rectangle (0.3,0.3);} Low Vegetation &
                    \raisebox{-0.5ex}{\tikz\fill[potsdam Tree] (0,0) rectangle (0.3,0.3);} Tree &
                    \raisebox{-0.5ex}{\tikz\fill[potsdam Car] (0,0) rectangle (0.3,0.3);} Car &
                    \raisebox{-0.5ex}{\tikz\fill[potsdam Clutter] (0,0) rectangle (0.3,0.3);} Clutter
                \end{tabular}
            };
        \end{tikzpicture}
        \caption{Vaihingen}
        \label{fig: Vaihingen}
    \end{subfigure}
    \caption{More visualization results of coarse-grained remote sensing images.}
    \label{fig: coarse-grained}
\end{figure*}

\begin{figure*}[t]
    \centering
    \begin{subfigure}[t]{0.9\textwidth}
        \centering
        \includegraphics[width=\textwidth]{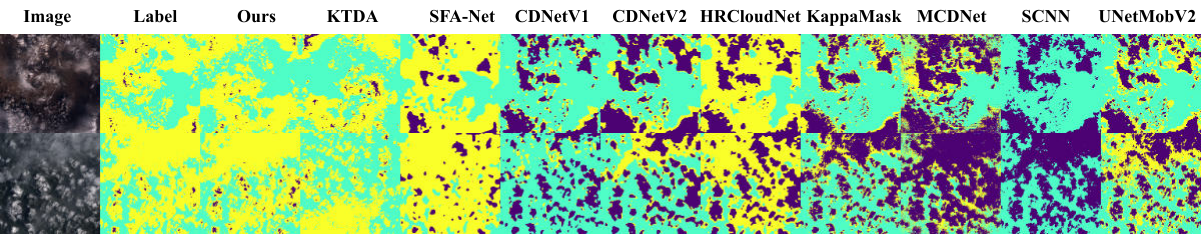}
        \vspace{10pt} 
        \begin{tikzpicture}[overlay, remember picture]
            \node at (0,0) {
                \fontsize{8}{8}\selectfont
                \begin{tabular}{*{4}{c@{\hspace{10pt}}}}
                    \raisebox{-0.5ex}{\tikz\fill[clear sky] (0,0) rectangle (0.3,0.3);} Clear Sky &
                    \raisebox{-0.5ex}{\tikz\fill[thick cloud] (0,0) rectangle (0.3,0.3);} Thick Cloud &
                    \raisebox{-0.5ex}{\tikz\fill[thin cloud] (0,0) rectangle (0.3,0.3);} Thin Cloud &
                    \raisebox{-0.5ex}{\tikz\fill[cloud shadow] (0,0) rectangle (0.3,0.3);} Cloud Shadow
                \end{tabular}
            };
        \end{tikzpicture}
        \caption{Cloud}
        \label{fig: Cloud}
    \end{subfigure}
    \hfill
    \begin{subfigure}[t]{0.9\textwidth}
        \centering
        \includegraphics[width=\textwidth]{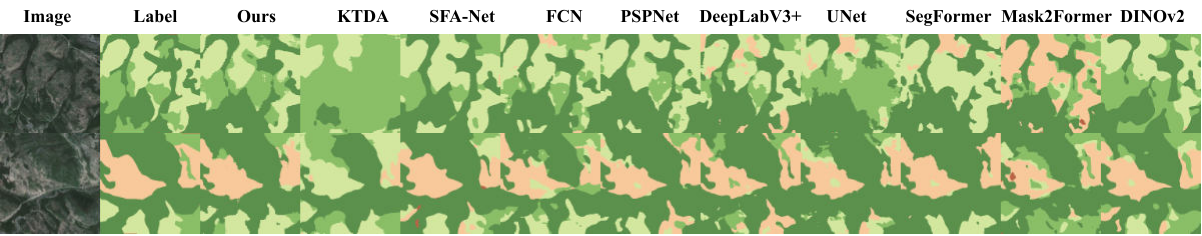}
        \vspace{10pt} 
        \begin{tikzpicture}[overlay, remember picture]
            \node at (0,0) {
                \fontsize{8}{8}\selectfont
                \begin{tabular}{*{5}{c@{\hspace{10pt}}}}
                    \raisebox{-0.5ex}{\tikz\fill[low] (0,0) rectangle (0.3,0.3);} Low &
                    \raisebox{-0.5ex}{\tikz\fill[middle low] (0,0) rectangle (0.3,0.3);} Middle-Low &
                    \raisebox{-0.5ex}{\tikz\fill[middle] (0,0) rectangle (0.3,0.3);} Middle &
                    \raisebox{-0.5ex}{\tikz\fill[middle high] (0,0) rectangle (0.3,0.3);} Middle-High &
                    \raisebox{-0.5ex}{\tikz\fill[high] (0,0) rectangle (0.3,0.3);} High
                \end{tabular}
            };
        \end{tikzpicture}
        \caption{Grass}
        \label{fig: Grass}
    \end{subfigure}
    \caption{More visualization results of fine-grained remote sensing images.}
    \label{fig: fine-grained}
\end{figure*}
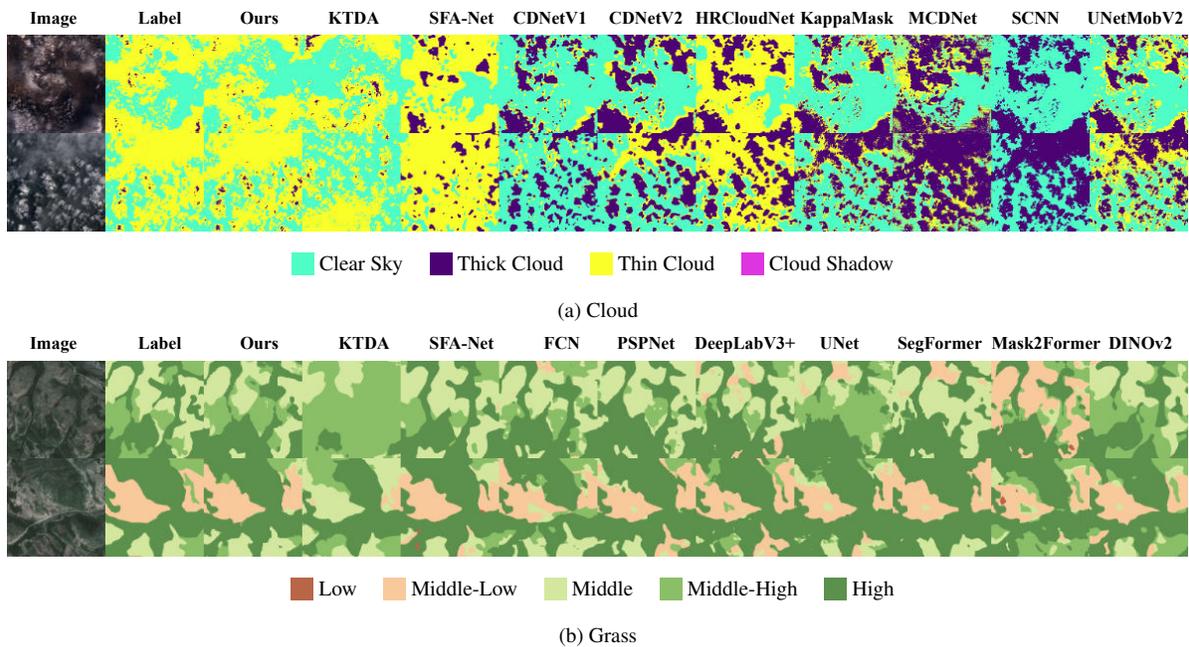

\begin{figure*}[t]
    \centering
    \includegraphics[width=0.9\textwidth]{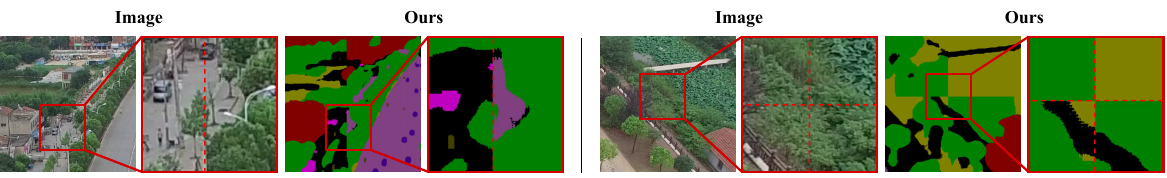} %
    \caption{Visualization examples of bad cases.}
    \label{fig: badcase}
\end{figure*}

\cref{fig: coarse-grained,fig: fine-grained} present comparative visualization results of our method against SFA-Net~\cite{sfanet} and UNetFormer~\cite{unetformer} across six challenging benchmarks. On the LoveDA (\cref{fig: LoveDA}) and UAVid (\cref{fig: UAVid}) datasets where test set ground truth is unavailable, our segmentation masks exhibit superior alignment with visual semantics compared to baselines, particularly in preserving structural continuity of buildings and road networks. The Potsdam (\cref{fig: Potsdam}) and Vaihingen (\cref{fig: Vaihingen}) results demonstrate our model's robustness against complex urban patterns, with significantly reduced segmentation artifacts in cluttered areas. Cloud (\cref{fig: Cloud}) and Grass (\cref{fig: Grass}) segmentation further validate our approach's capability to handle fine-grained texture variations, achieving state-of-the-art boundary precision. A failure mode analysis (\cref{fig: badcase}) reveals that tiling artifacts persist in UAVid inference due to our sliding window strategy, suggesting directions for future architectural improvements.

\section{Training Dynamics}~\label{sec: training_dynamics}

In this section, we present the curves showing how the metrics of our method and SFA-Net~\cite{sfanet} change with the training epochs. By comparing the performance across various aspects, we highlight the advantages of our method.

\paragraph{Training Stability}

\begin{figure*}[t]
\centering
\begin{minipage}[b]{0.3\linewidth}
  \centering
  \includegraphics[width=\linewidth]{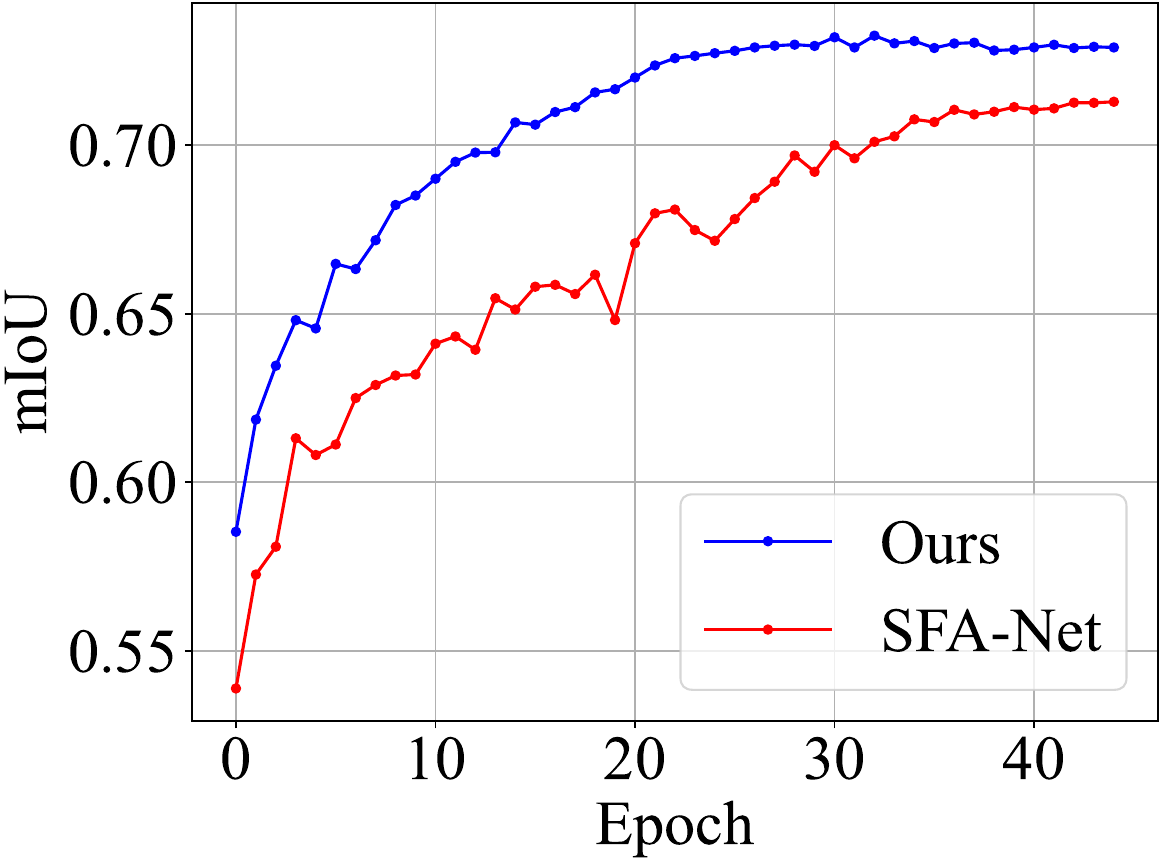}
  \subcaption{LoveDA}
  \label{fig:train_speed_a}
\end{minipage}
\hfill
\begin{minipage}[b]{0.3\linewidth}
  \centering
  \includegraphics[width=\linewidth]{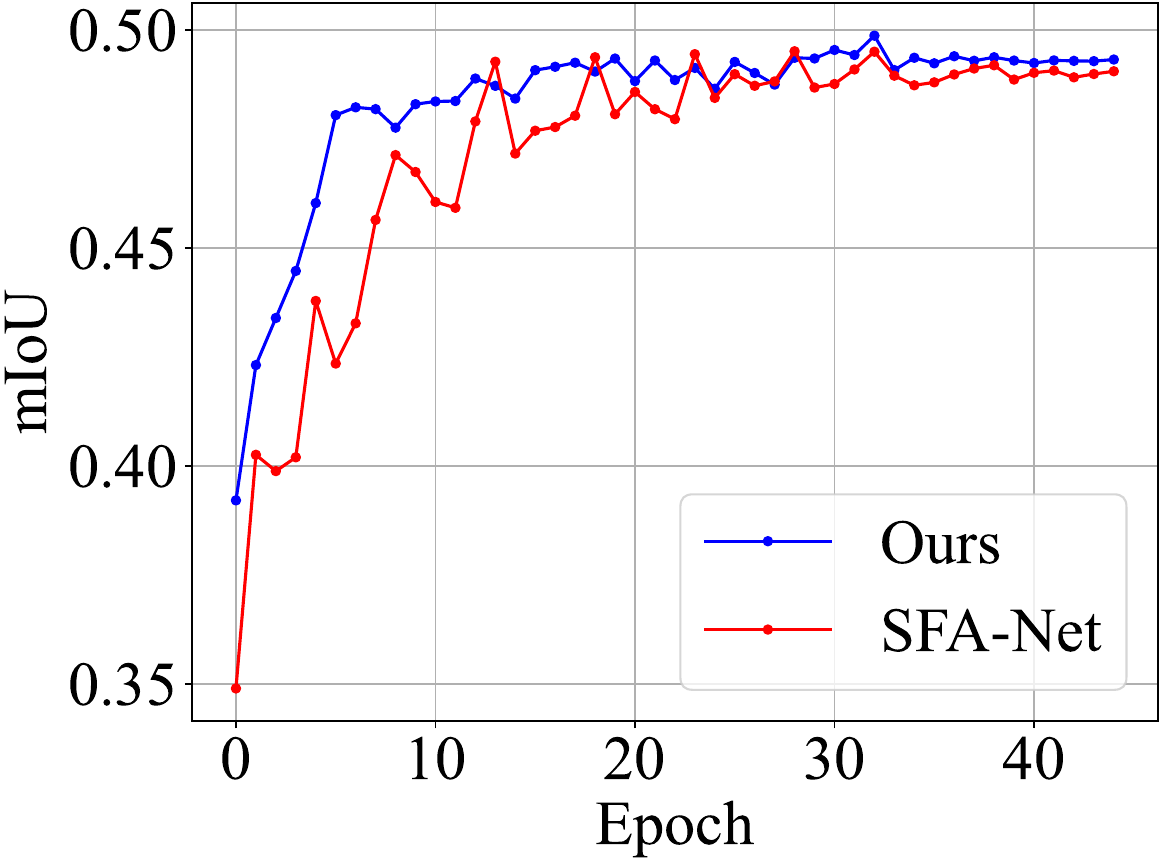}
  \subcaption{UAVid}
  \label{fig:train_stability_b}
\end{minipage}
\hfill
\begin{minipage}[b]{0.3\linewidth}
  \centering
  \includegraphics[width=\linewidth]{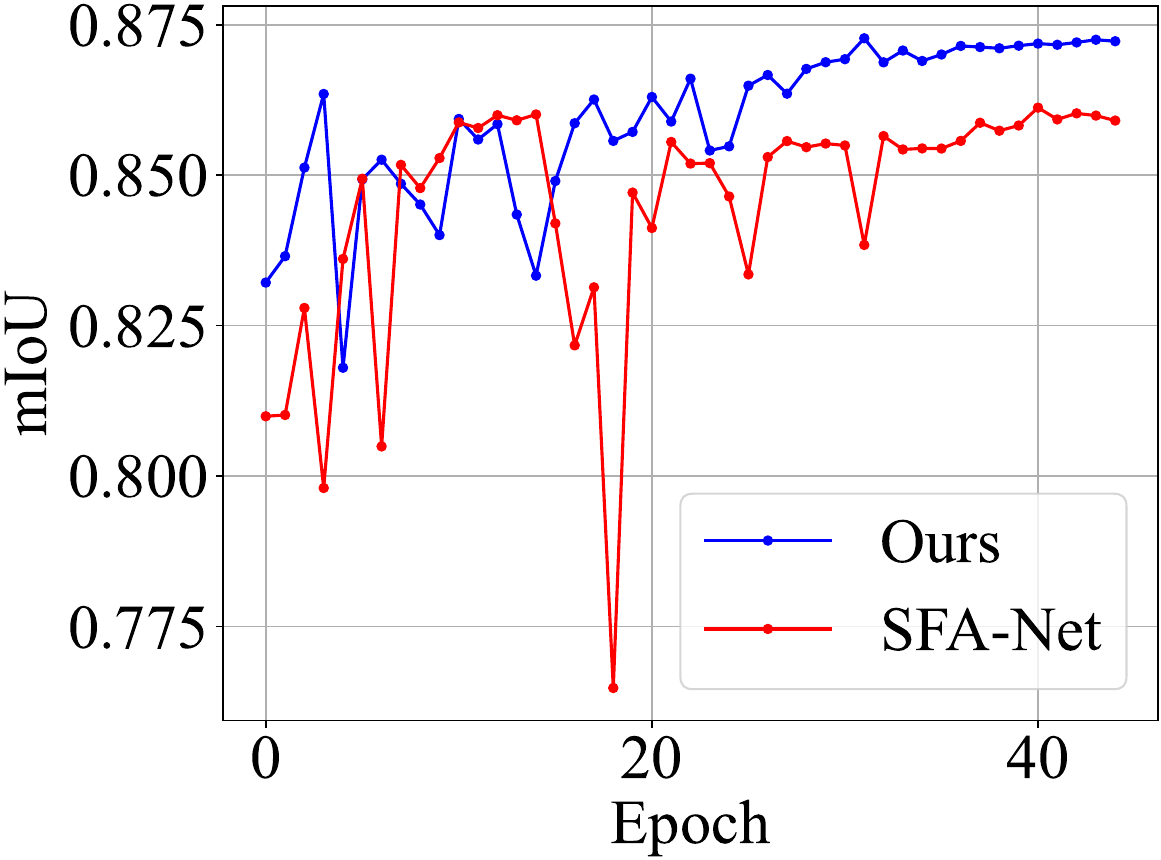}
  \subcaption{Potsdam}
  \label{fig:train_speed_c}
\end{minipage}

\vspace{1em} 

\begin{minipage}[b]{0.3\linewidth}
  \centering
  \includegraphics[width=\linewidth]{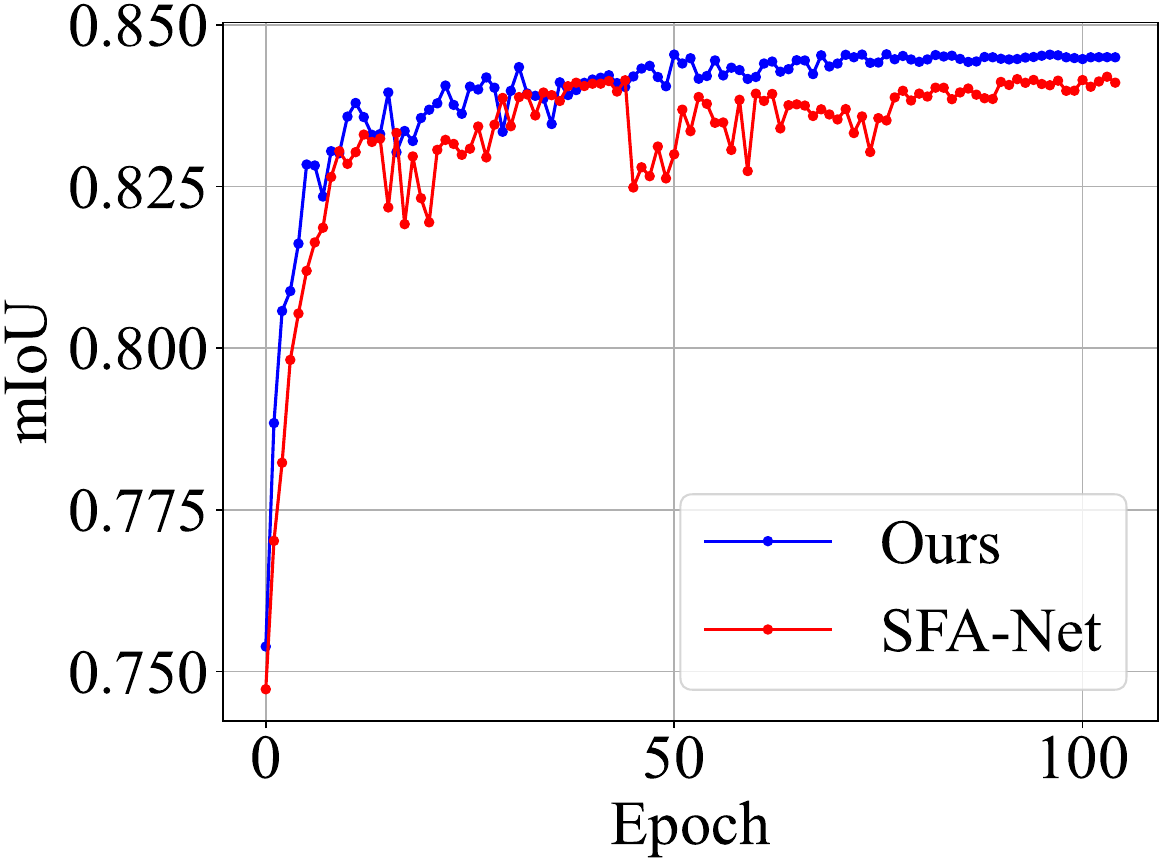}
  \subcaption{Vaihingen}
  \label{fig:train_speed_d}
\end{minipage}
\hfill
\begin{minipage}[b]{0.3\linewidth}
  \centering
  \includegraphics[width=\linewidth]{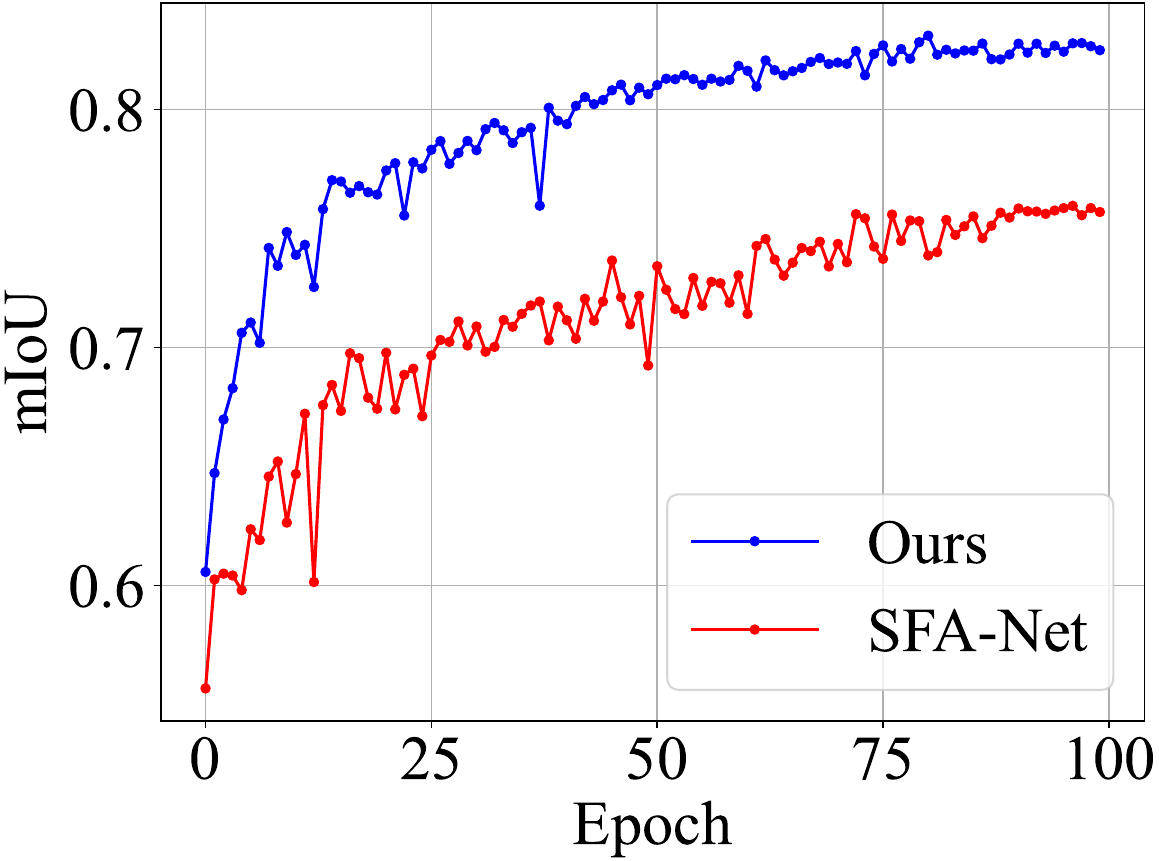}
  \subcaption{Cloud}
  \label{fig:train_performance_e}
\end{minipage}
\hfill
\begin{minipage}[b]{0.3\linewidth}
  \centering
  \includegraphics[width=\linewidth]{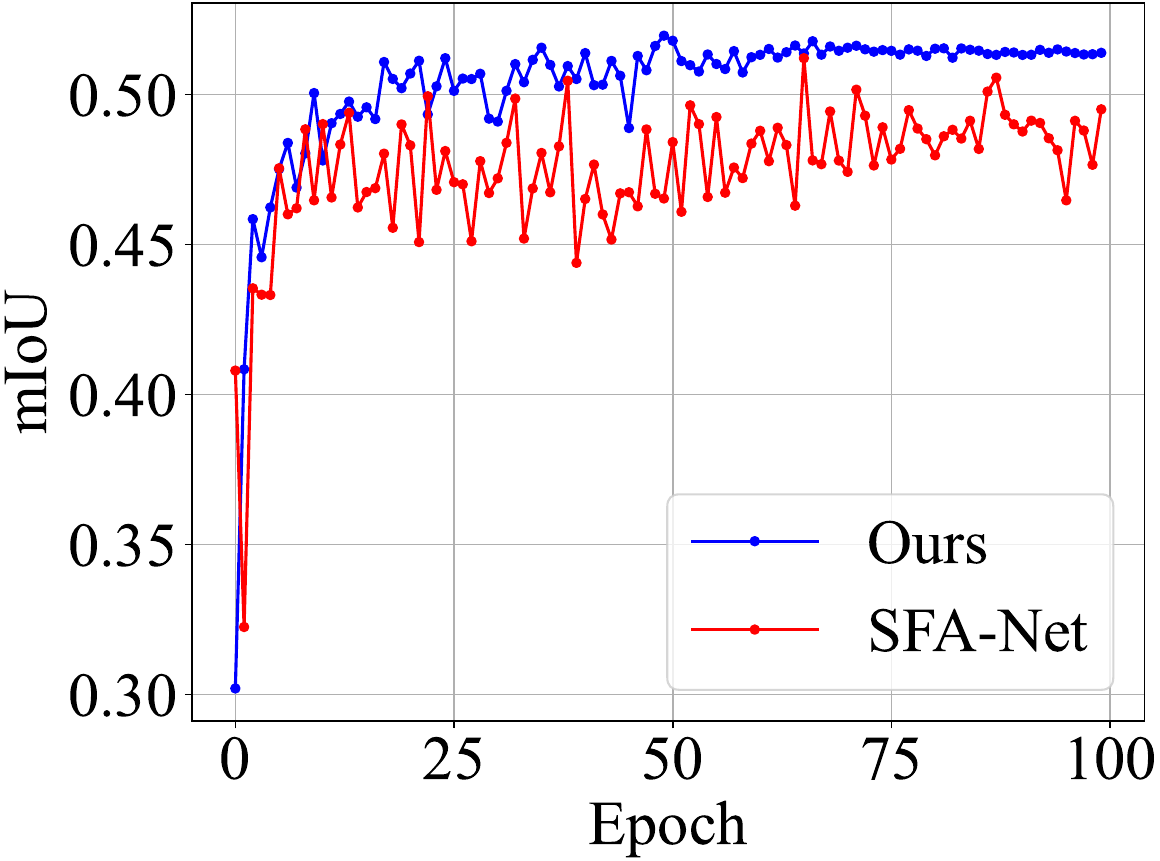}
  \subcaption{Grass}
  \label{fig:train_performance_f}
\end{minipage}

\caption{Validation set mIoU trends across all datasets over training epochs.}
\label{fig:all_experiments}
\end{figure*}

As shown in \cref{fig:train_stability_b} and \cref{fig:train_performance_f}, our method demonstrates superior stability during training. Compared to SFA-Net, our approach exhibits minimal fluctuations and maintains a steady convergence throughout the training process, which reduces the need for extensive hyperparameter tuning, saving computational resources.

\paragraph{Fitting Speed}

\cref{fig:train_speed_a} and \cref{fig:train_performance_e} illustrate the faster fitting speed of our method. Our model achieves optimal performance in fewer epochs, while SFA-Net requires more training time to reach similar results. This faster convergence enables more efficient model training.

\paragraph{Final Performance}

\cref{fig:all_experiments} holistically demonstrates our method's superiority across all evaluation metrics (subfigures a.-f.). The consistent performance advantages visible throughout the training lifecycle -- from initial convergence patterns to final stabilized outputs -- validate our approach's end-to-end effectiveness.

\end{document}